\documentclass{article}

\PassOptionsToPackage{square,numbers,sort&compress}{natbib}

\usepackage[utf8]{inputenc} 
\usepackage[T1]{fontenc}    
\usepackage[hidelinks]{hyperref}       
\usepackage{url}            
\usepackage{booktabs}       
\usepackage{amsfonts}       
\usepackage{nicefrac}       
\usepackage{microtype}      
\usepackage{xcolor}         

\usepackage{graphicx}
\usepackage{amsmath,mathtools,amsthm}
\usepackage{enumitem}
\graphicspath{{figs/}}
\usepackage{wrapfig}
\usepackage{algorithm}
\usepackage[noend]{algorithmic}

\usepackage{comment}
\usepackage[margin=1.27in]{geometry}

\usepackage{xr}

\makeatletter
\newcommand*{\addFileDependency}[1]{
  \typeout{(#1)}
  \@addtofilelist{#1}
  \IfFileExists{#1}{}{\typeout{No file #1.}}
}
\makeatother
\newtheorem{prop}{Proposition}

\usepackage{booktabs}
\newcommand{\ra}[1]{\renewcommand{\arraystretch}{#1}}

\usepackage{mathtools}
\mathtoolsset{showonlyrefs}

\newcommand{\indep}{\perp \!\!\! \perp}

\usepackage{bbm}

\newtheorem{theorem}{Theorem}

\newtheorem{corollary}{Corollary}
\newtheorem{remark}{Remark}

\graphicspath{{figs/}}
\usepackage{xr}
\usepackage{natbib}
\usepackage{booktabs}
\usepackage{caption}
\usepackage{subcaption}

\usepackage{mathtools}
\mathtoolsset{showonlyrefs}

\usepackage{bbm}

\usepackage{authblk}

\makeatletter
\def\@fnsymbol#1{\ensuremath{\ifcase#1\or \**\or \ddagger\or
   \mathsection\or \mathparagraph\or \|\or **\or \**\**
   \or \ddagger\ddagger \else\@ctrerr\fi}}
    \makeatother

\title{Learning to Increase the Power of Conditional Randomization Tests}

\date{}

\author[1]{Shalev Shaer \thanks{shalev.shaer@campus.technion.ac.il}}
\author[1,2]{Yaniv Romano \thanks{yromano@technion.ac.il}}
\affil[1]{Department of Electrical and Computer Engineering, Technion, Israel}
\affil[2]{Department of Computer Science, Technion, Israel}

\begin{document}

\maketitle

\begin{abstract}
The model-X conditional randomization test is a generic framework for conditional independence testing, unlocking new possibilities to discover features that are conditionally associated with a response of interest while controlling type-I error rates. An appealing advantage of this test is that it can work with any machine learning model to design powerful test statistics. In turn, the common practice in the model-X literature is to form a test statistic using machine learning models, trained to maximize predictive accuracy with the hope to attain a test with good power. However, the ideal goal here is to drive the model (during training) to maximize the power of the test, not merely the predictive accuracy. In this paper, we bridge this gap by introducing, for the first time, novel model-fitting schemes that are designed to explicitly improve the power of model-X tests.
This is done by introducing a new  cost function that aims at maximizing the test statistic used to measure violations of conditional independence.
Using synthetic and real data sets, we demonstrate that the combination of our proposed loss function with various base predictive models (lasso, elastic net, and deep neural networks) consistently increases the number of correct discoveries obtained, while maintaining type-I error rates under control.
    
\end{abstract}

\section{Introduction}
\label{sec:intro}

An important task in modern data analysis is identifying a subset of explanatory features from a larger pool that are associated with a response of interest. For instance, suppose a nutritionist is interested in identifying  which nutrient intakes are associated with body mass index. This information, in turn, could be leveraged to develop diets that improve human health \cite{chen2013variable}. As another example, consider a geneticist who is interested in finding which genetic mutations of a virus have developed resistance to a specific antiviral drug. To quote \cite{HIV}: ``[This understanding] is essential for designing new antiviral drugs and for using genotypic drug resistance testing to select optimal therapy.''
In such applications, there are often a large number of explanatory features with complex dependencies and a limited number of observations, which make the selection problem a particularly difficult task~\cite{shah2020hardness}. To stress this point, consider the genetic example from above and notice that a feature selection procedure can make erroneous selections of two kinds: (i) false positives---it may select mutations that do not have any effect on drug resistance; and (ii) false negatives---it may fail to discover genetic mutations that in fact developed resistance to the antiviral drug. This paper presents a novel cost function and learning scheme whose goal is to increase the number of discoveries reported by statistical methods for feature selection, while maintaining the rate of false positives under control.

To formalize the problem, we denote by $Y \in \mathbb{R}$ the response variable and by $X = (X_1,\dots,X_d) \in \mathbb{R}^d$ the feature vector. For example, imagine that $Y$ is a measure of drug resistance, and $X_j \in \mathbb{R}$---the $j$th entry in the vector $X$---indicates the presence or absence of a genetic mutation in a specific location. Conditional independence testing \cite{Knockoffs,HRT,dCRT,zhang2012kernel} deals with the question of whether a specific feature $X_j$ is independent of the response variable $Y\in \mathbb{R}$ after accounting for the effect of all the other features $X_{-j} = (X_1,\dots,X_{j-1}, X_{j+1},\dots,X_d) \in \mathbb{R}^{d-1}$. That is, the null hypothesis we seek to test is
\begin{equation} \label{eq:null}
    H_{0,j}: X_j \indep Y \mid X_{-j}
\end{equation}
against the alternative $H_{1,j}: X_j \not\!\indep Y \mid X_{-j}$. In plain words, $H_{0,j}$ inquires whether the knowledge of $X_j$ adds additional information about the response $Y$ beyond what is already contained in $X_{-j}$. Consequently, we say that the $j$th feature is null (i.e., unimportant) if $H_{0,j}$ is true, and non-null (i.e., important) if $H_{0,j}$ is false. 

Imagine we are given $n$ data points $\{(X^i,Y^i)\}_{i=1}^n$, where $X^i \in \mathbb{R}^d$ and $Y^i\in \mathbb{R}$ are drawn i.i.d. from a joint distribution $P_{XY}$. Statistical tests for conditional independence leverage the observed data to formulate a decision rule on whether to reject $H_{0,j}$ and report that the feature $X_j$ is likely to be important, or accept $H_{0,j}$ in cases where there is not enough evidence to reject the null. 
A test for conditional independence should return a \emph{valid} p-value $p$, which is a random variable satisfying $\mathbb{P}_{H_0}(p \leq \alpha) \leq \alpha, \  \alpha \in (0,1)$. This p-value is then used to rigorously control the type-I error rate, i.e., the probability of rejecting $H_{0,j}$ when it is in fact true, at any user-specified level $\alpha$, e.g., of 5\%. The power of the test (higher is better) is the true positive rate, being the probability of correctly reporting that the non-null $X_j$ is indeed important.

Thus far we formulated the problem of testing for a specific feature, however, in many scientific applications (e.g., genome wide association studies \cite{GenomCRT,benner2016finemap}) we are interested in selecting a subset of features that are associated with the response. In other words, we wish to test for all $H_{0,j},$ $j=1,\dots d$ \emph{simultaneously} while controlling some statistical notion of type-I error. To this end, imagine we have at hand $d$ p-values, one for each $H_{0,j}$, which we denote by $p_j$, $j=1, \dots ,d$.
Recall that each p-value allows us to control the type-I error for a specific hypothesis at a desired rate $\alpha$. However, when testing for all the hypotheses simultaneously we must account for the multiplicity of the tests, as otherwise the probability that some of the true null hypotheses are rejected by chance alone may be excessively high; see the survey in \cite{multiple_testing_correction} for details. Hence, we follow the seminal work reported in \cite{BH} and define our task as follows: identify \emph{the largest possible} subset $\hat{\mathcal{S}}\subseteq \{1,\dots,d\}$ of non-null features\footnote{{We assume there exists a unique Markov blanket $\mathcal{S}$, i.e., a unique smallest subset $\mathcal{S}$ such that $Y$ is independent of $\{X_j\}_{j \not\in \mathcal{S}}$ given $\{X_j\}_{j \in \mathcal{S}}$. We refer the reader to \cite[Section 2]{Knockoffs} for a detailed discussion in this regard.}} while ensuring that the false discovery rate (FDR), defined as
\begin{equation}
    \textrm{FDR} = \mathbb{E}\left[ \frac{|\hat{\mathcal{S}} \cap \mathcal{H}_0 |} {\textrm{max}\{|\hat{\mathcal{S}}|,1\}}\right],
\end{equation}
falls below a nominal level of choice $q\in (0,1)$, e.g., $q=0.1$ or $q=0.2$. Above, $\mathcal{H}_0 \subseteq \{1,\dots,d\} \setminus \mathcal{H}_1 $ contains the indices of the null features for which $H_{0,j}$ is true, $\mathcal{H}_1$ is the set of non-null features, and $|\cdot|$ returns the set-size. In words, the FDR is the expected proportion of true nulls among the rejected hypotheses. In our genetic example, the ability to control the FDR allows us to form a list of discoveries such that the majority of the selected mutations are expected to be conditionally associated with drug resistance, on average. This information may be valuable when allocating costly resources, required to further study the implications of the reported discoveries \cite{hawinkel2019broken, manolio2009finding}. The power of the selection procedure (larger is better) is defined as $
\mathbb{E}\left[  {| \hat{\mathcal{S}} \cap \mathcal{H}_1 |} / {|\mathcal{H}_1|}\right]$, being the expected proportion of non-nulls that are correctly selected among all the true non-nulls.
Controlling the FDR can be achieved by plugging the list of p-values $p_j,$ $j=1,\dots, d$ into the Benjamini-Hochberg procedure (BH)~\cite{BH}, as described in Section~\ref{sec:setup}. Importantly, the power of this feature selection procedure is affected by the ability of the underlying conditional independence test to generate small p-values for the non-null features, indicating strong evidence against the corresponding null hypotheses. Our goal in this work is to increase the power of the above controlled feature selection pipeline by forming data-driven test statistics that powerfully separate non-nulls from nulls. 

The model-X conditional randomization test \cite{Knockoffs} and its computationally efficient version---the holdout randomization test (HRT) \cite{HRT} that we study in this paper---are methods for conditional independence testing that gain a lot of attention in recent years. These tools are very attractive since they can leverage any powerful predictive model to rigorously test for $H_{0,j}$, under the assumption that the distribution of the explanatory features $P_X$ is known. As such, model-X tests are especially beneficial in situations where copious amount of \emph{unlabeled} data is available compared to labeled data, e.g., as happen in genome wide association studies \cite{GenomCRT,GWASFDR,geneticTwins}. 
{At a high level, the HRT is carried out by first splitting the data into a training set and a test set, and fit an arbitrary predictive model on the training set. Next, a dummy copy $\tilde{X}_j$ of the original feature $X_j$ is sampled for each sample in the test set, such that $\tilde{X}_j$ is independent of $Y$ given $X_{-j}$ by construction. In our running example, one can think of $\tilde{X}_j$ as a fake genetic mutation (sampled by a computer program) that does not carry any new information about $Y$ if we already know all the other mutations.
The test proceeds by comparing, for each observation in the test set, the accuracy of the trained predictive model applied to (i) the original feature vector $X$, and (ii) its modified version for which only $X_j$ is replaced by $\tilde{X}_j$. Here, a decrease in the overall accuracy can serve as an evidence that $X_j$ is associated with $Y$ after accounting for the effect of $X_{-j}$. }

In practice, the HRT is deployed by \emph{treating the machine learning model as a black-box}, fitted to maximize predictive accuracy with the hope to attain good power. However, the ultimate objective here is to maximize the power of the test and not merely the model's accuracy, and, currently, there is no method to balance between the two. To stress this point further, there is a growing evidence that modern machine learning algorithms tend to excessively rely on spurious features---null features that are solely correlated with the non-null ones---to make accurate predictions \cite{spurious1,spurious2,arjovsky2019invariant,peters2015causal,khani2021removing}. Loosely speaking, in such cases, the model tends to up-weight the importance of null features at the cost of down weighting the importance of the non-nulls. The consequence of this undesired phenomenon is that the predictive model may only show a mild drop in accuracy when replacing an important feature with its dummy copy. This renders the test to lose power, as the test accounts for the effect of the spurious correlations by design.

\subsection*{Our contribution}

We propose, for the first time, novel learning schemes to fit models that are designed to \emph{explicitly} increase the power of the HRT. This is done by `looking ahead' (during training) to what will happen when the predictive model is integrated within the HRT procedure. This stands in striking contrast with the common practice, in which the model is merely trained to maximize accuracy to the extent possible. 
Concretely, the core of our contribution is the formulation of a novel loss function that encourages the model's accuracy to drop when replacing the original feature $X_j$ with the dummy one $\tilde{X}_j$, seeking \emph{maximum risk discrepancy} (MRD). 
We introduce in Section~\ref{sec:proposed} a general, theoretically motivated, stochastic optimization procedure that enables us to augment our MRD loss to existing cost functions used in the variable selection literature, e.g., sparsity promoting penalty for deep neural networks.
That section also includes a specialized learning scheme to fit sparse linear regression models (such as lasso) with our proposed loss. Extensive experiments on synthetic and real data sets are presented in Section~\ref{sec:experiments}, showing that the combination of the proposed MRD approach with existing regression algorithms \emph{consistently improves} the power of the underlying controlled feature selection procedure. {In particular, in cases where the signal is weak (i.e., in low power regimes), our method may even achieve more than 90\% relative improvement compared to existing methods.} Importantly, the MRD scheme preserves the validity guarantee of the p-values generated by HRT. 
Furthermore, we demonstrate that our new proposal is fairly robust to situations where the distribution of $P_X$---required to sample the dummy features---is estimated from data. These situations are of course ubiquitous in real-world applications. 
Lastly, a software package that implements our methods is available at \url{https://github.com/shaersh/MRD}.

\section{Background}

\subsection{Model-X randomization test}
\label{model_x}

The model-X randomization test provides a valid p-value for $H_{0,j}$ in finite samples without making modeling assumptions on the conditional distribution of $Y \mid X$. However, this approach assumes that the conditional distribution of $X_j \mid X_{-j}$ is known.
The original conditional randomization test (CRT), developed in \cite{Knockoffs}, requires fitting a regression model many times in order to be carried out precisely. The HRT, which is extensively used in this paper, 
bypasses this computational issue at the expense of data splitting. Specifically, this test starts by dividing the data into disjoint training and testing sets. The training set is used to fit an arbitrary regression model $\hat{f}$, formulating a test statistic function $T(X_j, X_{-j},Y;\hat{f}) \in \mathbb{R}$, e.g., the model's prediction error.
Let $\{(X_i,Y_i)\}_{i\in\mathcal{I}}$ be the samples of the testing set. The HRT proceeds by repeating the following two steps for each $k=1,\dots,K$ (treating $\hat{f}$ as a fixed function):
\begin{itemize}
    \item Sample a dummy feature $\tilde{X}^i_j \sim P_{X_j|X_{-j}}(X_j^i|X_{-j}^i)$, for all samples in the test set $i\in\mathcal{I}$.
    \item Compute the test statistic using the dummy variables $$\tilde{t}_j^{(k)} \gets \frac{1}{|\mathcal{I}|} \sum_{i\in\mathcal{I}} T(\tilde{X}_{j}^i,X_{-j}^i,Y^i;\hat{f}).$$
\end{itemize}
Finally, a p-value $\hat{p}_j$ for $H_{0,j}$ is constructed by computing the empirical quantile {level} of the true test statistic $t^* \gets \frac{1}{|\mathcal{I}|} \sum_{i\in\mathcal{I}} T(X_j^i, X_{-j}^i,Y^i;\hat{f})$ among the dummy statistics $\tilde{t}_j^{(1)},\dots,\tilde{t}_j^{(K)}$:
\begin{equation} \label{eq:p-value}
    \hat{p}_j=\frac{1+ \left| \{ k : t^* \geq \tilde{t}_j^{(k)} \}\right|}{{K+1}}.
\end{equation}
Importantly, the p-value $\hat{p}_j$ in \eqref{eq:p-value} is valid, i.e., $\mathbb{P}[\hat{p}_j \leq \alpha \mid H_{0,j} \ \text{is true}] \leq \alpha$. This is because (i) $\tilde{X}_j \indep Y \mid X_{-j}$ by construction, and (ii) under the null, the random variables $t^*, \tilde{t}_j^{(1)}, \dots, \tilde{t}_j^{(K)}$ are exchangeable \cite{Knockoffs}.

The validity of the p-value holds true for any choice of predictive model $\hat{f}$, and for any choice of test statistic $T(\cdot)$ used to compare the distributions of $(Y, {X}_j, X_{-j})$ and $(Y, \tilde{X}_j, X_{-j})$. The authors of \cite{TheoCRT} proved that the likelihood ratio is the most powerful statistic, however, it requires the knowledge of the true conditional distribution of $Y \mid X$, which is unknown. In practice, a common choice for a test statistic---which we use in our proposal as well---is the squared error \cite{HRT,geneticTwins}:
\begin{equation}
    T(X_j, X_{-j},Y;f)=(Y-\hat{f}(X_j, X_{-j}) )^2,
    \label{HRT_tstat}
\end{equation}
where we use the notation $\hat{f}(X_j, X_{-j}) = \hat{f}(X)$ to stress that $\hat{f}$ is applied on the vector $X$ with its original $j$th feature $X_j$, whereas $\hat{f}(\tilde{X}_j, X_{-j})$ is applied to the same feature vector except that the original $X_j$ is replaced by its dummy copy $\tilde{X}_j$. To handle the controlled feature selection problem, the HRT p-values $p_j$, $j=1,\dots,d$ are most often plugged into the Benjamini-Hochberg procedure (BH)~\cite{BH} procedure, as discussed next.

\subsection{FDR control via the BH procedure} \label{sec:setup}

Armed with a list of $d$ p-values $p_j$, one for each $H_{0,j}$, we can apply the BH procedure~\cite{BH} to control the FDR. This method operates as follows. First, sort the given $d$ p-values in a non-decreasing order; we denote by $p_{(k)}$ and $H_{0,(k)}$ the $k$th smallest p-value and its corresponding null-hypothesis, respectively. Then, for a given FDR target level $q$, find the largest $k_0$ such that $p_{(k_0)} \leq \frac{k_0}{d}q$. Lastly, reject the null hypotheses $H_{0,(1)}, H_{0,(2)}, \dots, H_{0,(k_0)}$. Importantly, the procedure described above is guaranteed to control the FDR under the assumption that the p-values are independent. While the p-values we construct in this work do not necessarily satisfy the independence assumption, it is known that the BH procedure often controls the FDR empirically even when the p-values are dependent (except adversarial cases). This empirical observation is also endorsed in our experiments, standing in line with many other methods that plug the dependent p-values generated by CRT/HRT into the BH procedure \cite{HRT,Knockoffs}.
With that said, it is worth noting that there exists a variant of the BH that provably controls the FDR for any arbitrary dependency structure of $p_j,$ $j=1,\dots,d$, however, this comes at the cost of reduced power~\cite{BY}.

\section{The proposed method}
\label{sec:proposed}

In this section, we introduce our MRD approach to fit predictive models with the goal of explicitly improving the power of the test. We begin by describing the theoretical motivation behind our proposal. Then, we present a general learning scheme to fit an arbitrary predictive model (e.g., a neural network) using the MRD loss. Lastly, we design a specialized learning scheme for sparse linear regression problems.

\subsection{Main idea}
A central definition in our proposal is that of the risk discrepancy, expressed as
\begin{equation}\label{eq:population_RD}
    \textrm{RD}_j = \mathbb{E}[T(Y,\tilde{X}_j,X_{-j}; \hat{f}) - T(Y,X_j,X_{-j}; \hat{f})],
\end{equation}
where $\hat{f}$ is a fixed (pre-trained) predictive model, and $T(\cdot)$ is the squared error~\eqref{HRT_tstat}.
In the context of the HRT (Section~\ref{model_x}), a test with good power would have large and positive empirical risk discrepancy for the non-null features ${j \in \mathcal{H}_1}$.
Therefore, during training, we wish to encourage the model $\hat{f}$ to maximize $\textrm{RD}_j$ \emph{only for that group of features}. While this is infeasible because we do not have apriori knowledge of which of the features are in $\mathcal{H}_1$, we do have access to the sampling distribution of the conditional null through the sampling of the dummy copies $\tilde{X}_j$. This property is formally presented in the following well-known result in the model-X literature \cite{Knockoffs}, which we will use later to formulate our loss.
\begin{prop} [Cand\`{e}s et al. 2018 \cite{Knockoffs}]
Take $(X,Y) \sim P_{XY}$, and let $\tilde{X}_j$ be drawn independently from $P_{X_j \mid X_{-j}}$ without looking at $Y$. If $Y \indep X_j \mid X_{-j}$, then $(Y,\tilde{X}_j,X_{-j}) \overset{d}{=} (Y,{X}_j,X_{-j})$.
\label{prop1}
\end{prop}
\noindent Above, the notation $\overset{d}{=}$ stands for equality in distribution. For completeness, the proof of the above statement is given in Supplementary Section~\ref{supp:proofs}.
A consequence of Proposition~\ref{prop1}, important to our purposes, is given in the next corollary; see also \cite{RFI}.
\begin{corollary}
Let $\hat{f}$ be a fixed predictive model, and $T(\cdot; \hat{f})$ be any test statistic.
Under the setting of Proposition~\ref{prop1}, if $(Y,\tilde{X}_j,X_{-j}) \overset{d}{=} (Y,{X}_j,X_{-j})$, then $\mathbb{E}[T(Y,\tilde{X}_j,X_{-j}; \hat{f}) - T(Y,X_j,X_{-j}; \hat{f})]=0$.
\label{corollary1}
\end{corollary}
\noindent This motivates our formulation of a training procedure that drives the model to a solution in which the empirical risk discrepancy (evaluated on the training data) is maximized for all features, knowing that this quantity is guaranteed to be small for the null features at test time. In fact, in Proposition~\ref{prop2}, which is presented formally in Section \ref{sec:linear_mrd}, we prove that for a linear model with uncorrelated features, the solution that minimizes the population version of our objective function would result in estimated regression coefficients that are guaranteed to be zero for the null features.

\subsection{A general learning scheme}
\label{sec:gen_scheme}
\begin{algorithm}[t]
\textbf{Input}: Training data $\{(X^i,Y^i)\}_{i=1}^m$; predictive model $f_\theta$ with parameters $\theta$; step size $\eta$; baseline objective hyperparameters; MRD hyperparameter $\lambda$;  number of features to generate $1 \leq N \leq d$.

\begin{algorithmic}[1]
\FOR{$k = 1,\dots,K$}
\STATE $\textrm{Sample a random subset of features } \mathcal{P} \subseteq \{1 \cdots\ d\} \textrm{ of the size } N.$

\STATE $\textrm{Generate a dummy }\tilde{X}_j^i \textrm{ for each training point } X_i \textrm{ and for each } j \in \mathcal{P}$.

\STATE Update the model parameters $\theta$ using stochastic gradient descent:
$$
\theta^{k+1} \leftarrow \theta^k - \eta \nabla_\theta \bigg[ (1-\lambda) \mathcal{J}_{\mathrm{base}}(\textbf{Y},\textbf{X};f_\theta) +\frac{\lambda}{|\mathcal{P}|} \sum_{j \in \mathcal{P}} \mathcal{D} (z(f_\theta),\tilde{z}(f_\theta)) \bigg],
$$

where $z(f_\theta),\tilde{z}(f_\theta)$ stress that $z, \tilde{z}$ are functions of $\theta$.
\ENDFOR
\end{algorithmic}
\vspace{0.1cm}
\textbf{Output}: A fitted model $\hat{f}$.
\caption{Fitting an MRD model}
\label{alg:general_optimization}
\end{algorithm}

We now formalize the idea presented above. Given a training set of $m$ i.i.d samples $\{(X^i,Y^i)\}_{i=1}^m$, our procedure begins with a generation of a dummy copy $\tilde{X}^i_j$ for each sample and feature:
\begin{equation*}
\tilde{X}^i_j \sim P_{X_j \mid X_{-j}}(X^i_j \mid X^i_{-j}), \quad i=1,\dots,m, \ \  j = 1 \dots d.
\label{synthetic}
\end{equation*}
Denote the collection of training features by $\textbf{X} \in \mathbb{R}^{m \times d}$, where $X^i$ is the $i$th row in that matrix, and by $\textbf{Y} \in \mathbb{R}^{m}$ a vector that contains the response variables $Y^i$ as its entries. Let $\mathcal{J}_{\textrm{base}}(\textbf{Y},\textbf{X}; f)$ be the objective function of the base predictive model, expressed as
\begin{equation}
\mathcal{J}_{\textrm{base}}(\textbf{Y},\textbf{X}; f) = \frac{1}{m} \sum_{i=1}^m \ell(f(X^i),Y^i) + \alpha\mathcal{R}(f).
\label{baseline_objective}
\end{equation}
Above, the loss function $\ell$ measures the prediction error (MSE in our experiments), $\mathcal{R}$ is a regularization term, and the hyperparameter $\alpha$ trades off the importance of the two. With this notation in place, our proposal aims to minimize the following cost function: 
\begin{equation}
\hat{f}(x) = \underset{f\in \mathcal{F}}{\textrm{argmin}} \ (1-\lambda) \mathcal{J}_{\textrm{base}}(\textbf{Y},\textbf{X}; f) + \frac{\lambda}{d} \sum_{j=1}^{d} {\mathcal{D}( z, \tilde{z}_j)},
\label{mrd_objective}
\end{equation}
where $z=\frac{1}{m} \sum_{i=1}^m {T(Y^i,X_j^i,X_{-j}^i;f)}$ and  $\tilde{z}_j=\frac{1}{m} \sum_{i=1}^m {T(Y^i,\tilde{X}_j^i,X_{-j}^i;f)}$ are the empirical risks{---MSE in our context since $T(\cdot)$ is the squared error~\eqref{HRT_tstat}---}of the model evaluated on the original and synthetic triplets, respectively.
The function $\mathcal{D}(z, \tilde{z}_j)\in \mathbb{R}$ measures the difference between $z$ and $\tilde{z}_j$, such that $\mathcal{D}(z, \tilde{z}_j)$ gets smaller as $ \tilde{z}_j$ becomes larger than $z$. Therefore, \eqref{mrd_objective} seeks a model $\hat{f}$ that aligns with the base objective $\mathcal{J}_{\textrm{base}}$ while creating a better separation between $z$ and $\tilde{z}_j$. The parameter $\lambda \in [0,1]$ sets the relative weight between the two functionals; we present a simple, automatic approach for tuning this parameter in Supplementary Section~\ref{supp:models}.  
In practice, we define $\mathcal{D}$ as follows: $\mathcal{D}(z,\tilde{z}_j)=\sigma(z-\tilde{z}_j),$ where $\sigma(\cdot)$ is the Sigmoid function, i.e., $\sigma(s)=1/(1+e^{-s})$.
Of course, other forms of this function, such as the ratio $z / \tilde{z}_j $, can be considered. 

For ease of reference, Algorithm~\ref{alg:general_optimization} presents a general approach for minimizing~\eqref{mrd_objective}. In short, we use stochastic optimization and back-propagation, where each gradient step updates the model parameters by resampling $\tilde{X}_j$ for a subset of features from $\{1,\dots,d\}$. {We sample new dummies at each iteration to reduce algorithmic randomness, which, in turn, increases the stability of the procedure}. In Supplementary Section~\ref{supp:optim} we provide a weak form of convergence of this algorithm, where the analysis requires $\mathcal{D}(\cdot)$ to be bounded from below. A requirement that is satisfied by our choice to use the Sigmoid function. 
Algorithm~\ref{alg:general_optimization} is used in our experiments, implemented with a neural network as the base predictive model. 

\begin{remark}
Algorithm~\ref{alg:general_optimization} implements the RD penalty for all features simultaneously, thereby avoiding the need to fit a different model for each feature separately. This algorithm can be modified to fit a model with an RD penalty defined for any subset of features, and in particular to a specific feature $X_j$, as we demonstrate in Supplementary Section~\ref{sec:single_hyp}.
\end{remark}

\subsection{A scheme for sparse linear regression}
\label{sec:linear_mrd}
Next, we offer an optimization procedure tailored for the choice of sparse linear regression as the base objective, since it is widely used in the feature selection literature and often results in a powerful test~\cite{Knockoffs_original,HRT}. 
To this end, consider the combination of the elastic net objective~\cite{eNet} with our MRD loss: 
\begin{equation*}
\label{sparse_reg_objective_MRD}
\begin{array}{ll}
    \hat{\beta} = \underset{\beta}{\textrm{argmin}} \  (1-\lambda)\bigg(\underbrace{\frac{1}{2m} || \textbf{X}\beta - \textbf{Y}||^2_2 + \alpha_1 ||\beta||_1 + \frac{\alpha_2}{2}||\beta||_2^2}_{\mathcal{J}_{\textrm{base}}(\textbf{Y},\textbf{X}; f)}\bigg) + \frac{\lambda}{d} \sum_{j=1}^{d} {\mathcal{D}( z(\beta), \tilde{z}_j(\beta))}, 
\end{array}
\end{equation*}
where $\beta \in \mathbb{R}^d$ is the regression coefficient vector,
and $\alpha_1,\alpha_2$ are the elastic net penalty parameters. Above, we use the notations $z(\beta)$ and $ \tilde{z}_j(\beta)$ to stress that $z$ and $ \tilde{z}_j$ are functions of $\beta$.
Using variable splitting, we can separate the sparsity promoting penalty from the remaining terms, and solve the resulting optimization problem using the alternating direction method of multipliers ADMM \cite{ADMM}; see Supplementary Section~\ref{supp:sparse} for more details. Following Supplementary Algorithm~\ref{alg:admm}, the advantage of the above splitting strategy is that the minimization with respect to $\beta$ has a closed-form solution, being a simple projection onto the elastic net ball~\cite{enet_proj,enet_proxi}. In practice, we run this iterative algorithm until the stopping criteria suggested by \cite[Section 3.3]{ADMM} is met.

We conclude this section by showing that under the assumptions of Proposition~\ref{prop2}, the estimated coefficients of the null features $\hat{\beta}_j, j \in \mathcal{H}_0$ obtained by minimizing the MRD objective will be equal to zero. The proof is provided in Supplementary Section~\ref{supp:proofs}.
\begin{prop}
Suppose that the population model is linear $Y=X^T\beta+\epsilon$, where $\mathbb{E}[X_jX_k]=0$ for $1 \leq k \neq j \leq d$, and $\mathbb{E}[\epsilon X_j] = 0 $ for all $1\leq j \leq d$. Denote by $\hat{\beta}$ a solution to the infinite-data version of the MRD:
\begin{align}
     &\hat{\beta} = \underset{\beta}{\textup{argmin}} \ (1-\lambda)\mathbb{E}{[(X^T\beta - Y)^2]} +\frac{\lambda}{d} \sum_{j=1}^d {\sigma(\mathbb{E}[T(Y,\tilde{X}_j,X_{-j}; \beta) - T(Y,X_j,X_{-j}; \beta)])}. 
\end{align}
Under the above assumptions, for any $0 \leq \lambda < 1$, and for all $j \in \mathcal{H}_0$ with $\beta_j=0$, the solution $\hat{\beta}$ satisfies that $\hat{\beta}_j = 0$ for all $j \in \mathcal{H}_0$. 
\label{prop2}
\end{prop}

\section{Related work}
\label{sec:related}

\subsection{Handling unknown conditionals}
\label{sec:unknown_p}
Our proposed method inherits the model X assumption that the marginal distribution $P_X$---required to sample $\tilde{X}_j$---is known: this assumption is needed to guarantee that \eqref{eq:p-value} is a valid p-value \cite{Knockoffs}. However, in most real-world applications $P_X$ is unknown and thus must be estimated from the data. Here, dummy features of poor quality may break the validity of model-X randomization tests \cite{DeepKnockoffs, HRT, Contra}, including our proposal. To alleviate this concern, model X tests are equipped with various methodologies to estimate $P_X$, including a second-moment multivariate Gaussian estimator \cite{Knockoffs}, mixture density networks  \cite{HRT}, generative adversarial networks \cite{GANCRT}, moment matching networks \cite{DeepKnockoffs}, and more \cite{sesia2019gene,sudarshan2020deep, gimenez2019knockoffs,Contra}.
Furthermore, model-X tests are also equipped with diagnostic tools to improve the confidence in the results obtained by this method when the conditionals are unknown. These tools include goodness-of-fit tests, designed to quantify the quality of the generated dummies, as well as controlled semi-synthetic experiments with simulated response variables that allow the user to compare the obtained empirical FDR with the desired level \cite[Section~5]{DeepKnockoffs}.

Importantly, empirical evidence show that the model-X tests are robust in the sense that the FDR is often controlled in practice, especially when the testing procedure is applied with well-approximated dummy features. This is also corroborated by our experiments on real and simulated data: we practically obtain FDR control even when using the simplest Gaussian approximation to the unknown~$P_X$, highlighting the robustness to model misspecification of the HRT in general, and our proposal, in particular.

\subsection{Relevant prior art}

Our proposal is connected to the two-stage ``learn-then-test'' approach, initiated by the influential work of~\cite{optimal_2sample} in the context of kernel-based tests. The original idea here is to optimize the kernel parameters in order to maximize the power of the underlying test, where ample evidence shows that this approach is extremely fruitful \cite{optimal_2sample,kernelized_discrepancy,kernel_to_fit}. These contributions are in line with our proposal, although the latter is different as it is tailored to work in combination with the HRT.

Recently, new deep learning methods for feature selection have been proposed to improve data analysis with non-linear interactions. In particular, \cite{CancelOut,LassoNet,stochastic_gates} offer sparsity-promoting layers that can be combined with arbitrary neural network architectures. These techniques can be combined with our MRD framework, as demonstrated in Section~\ref{sec:experiments}, where we utilize the CancelOut regularizer developed in~\cite{CancelOut}. A different but related line of work is that of instance-wise interpretability methods, aiming at identifying a subset of relevant features for a single data point \cite{l2x,invase}.
While this line of work is outside the scope of this paper, we believe these methods may benefit from our MRD approach. {For instance, by combining the MRD with the test proposed by \cite{burns2020interpreting}, which is capable of finding instance-wise
features that a given predictive model considers important.}

\section{Experiments}
\label{sec:experiments}

In this section, we study the performance of the proposed MRD approach in a wide variety of settings. These include experiments with linear and nonlinear simulated data of varying signal strength and sample size, model misspecification experiments, and similar experiments with semi-synthetic and real data. The conclusion of this comprehensive study is that the MRD approach \emph{consistently} improves the power of HRT, more significantly in low power~regimes.

\subsection{Synthetic experiments}
\label{sec:syn_exp}

To evaluate the performance of our methods, we implement variable selection in a fully controlled synthetic setting in which the distribution of $X$ and $Y \mid X$ are known. This experimental setting allows us to study the validity and statistical efficiency of a wide variety of predictive models $\hat{f}$ integrated with the HRT.
Following \cite{Knockoffs,DeepKnockoffs,DeepPink}, we generate $X\in \mathbb{R}^d$ of dimension $d=100$ from a multivariate Gaussian distribution $\mathcal{N}(0,\Sigma)$ with $\Sigma_{i,j}=\rho^{|i-j|}$ and an auto-correlation parameter $\rho$. The response $Y \in \mathbb{R}$ is sampled from a sparse regression model of the form
\begin{equation} \label{eq:y-mid-x}
    Y = g(X) + \epsilon,
\end{equation}
where $\epsilon \sim \mathcal{N}(0,1)$ is the noise component and $g$ is a link function. Throughout the synthetic experiments we consider the following conditional models of $ Y \mid X$:
\begin{itemize}
    \item \textbf{M1: linear model}, with $g(X)= X^{\top}\beta$, where $\beta \in \mathbb{R}^d$ has $30\%$ randomly chosen non-zero entries of a random sign and a magnitude equal to $c$.
    \item \textbf{M2: polynomial model}, we follow \cite{DeepPink} and set a nonlinear $g(X)= (X^{\top}\beta)^3/2$, where $\beta$ is chosen as described in \textbf{M1}.
    \item \textbf{M3: sum of sines} is another nonlinear model we explore, with $g(X) = \sum_{j \in \mathcal{H}_1} {\sin{(X_j\beta_j)}}$. Here, $X_j$ and $\beta_j$ are the $j$th entries of $X$ and $\beta$, respectively; $\beta$ is defined as in \textbf{M1}.
    \item \textbf{M4: interaction model}, a challenging nonlinear model with $g(X) = \sum_{j=1}^{15} { X_{2j}X_{2j-1} }$.
\end{itemize}
Note that for data models \textbf{M1}-\textbf{M3}, a variable $X_j$ is in the non-null set $j \in \mathcal{H}_1$ if and only if $\{j=1,\dots,d : \beta_j \neq 0\}$. In the same vein, \textbf{M4} has 30 non-null variables, those who appear in the function $g(x)$. 

\noindent To apply the test, we first draw $m$ training examples from $P_{XY}$, which are used to fit a predictive model of choice. We then generate an additional independent set of $m$ test points, which are used to compute a p-value for each of the hypotheses $H_{0,j}, j=1,\dots,d$.
We use lasso, elastic net, and neural network as baseline models, where the latter is deployed only in the non-linear setting. See Supplementary Section~\ref{supp:models} for additional information on the training strategy and choice of hyperparameters. We apply our MRD algorithm in combination with the above methods using the same choice of hyperparameters. In addition, we set the RD penalty parameter $\lambda$ to be proportional to the validation error of the base predictive model. Supplementary Section~\ref{supp:models} describes how we evaluate this error for each base model. All models are fit by normalizing the features and response variables to have a zero mean and unit variance.

\subsubsection{An illustrative example}
\label{sec:illustrative_example}

We begin with a synthetic experiment that demonstrates the effect of our proposed training scheme on the power of the selection procedure.
To this end, we use $m=400$ samples for training, fix the auto-correlation parameter $\rho$ to be $0.1$, and construct $ Y \mid X$ that follows a polynomial data generating function (\textbf{M2}) with a signal amplitude $c=0.14$.

\begin{figure}[t]
  \centering
    \includegraphics[width=0.9\textwidth]{ 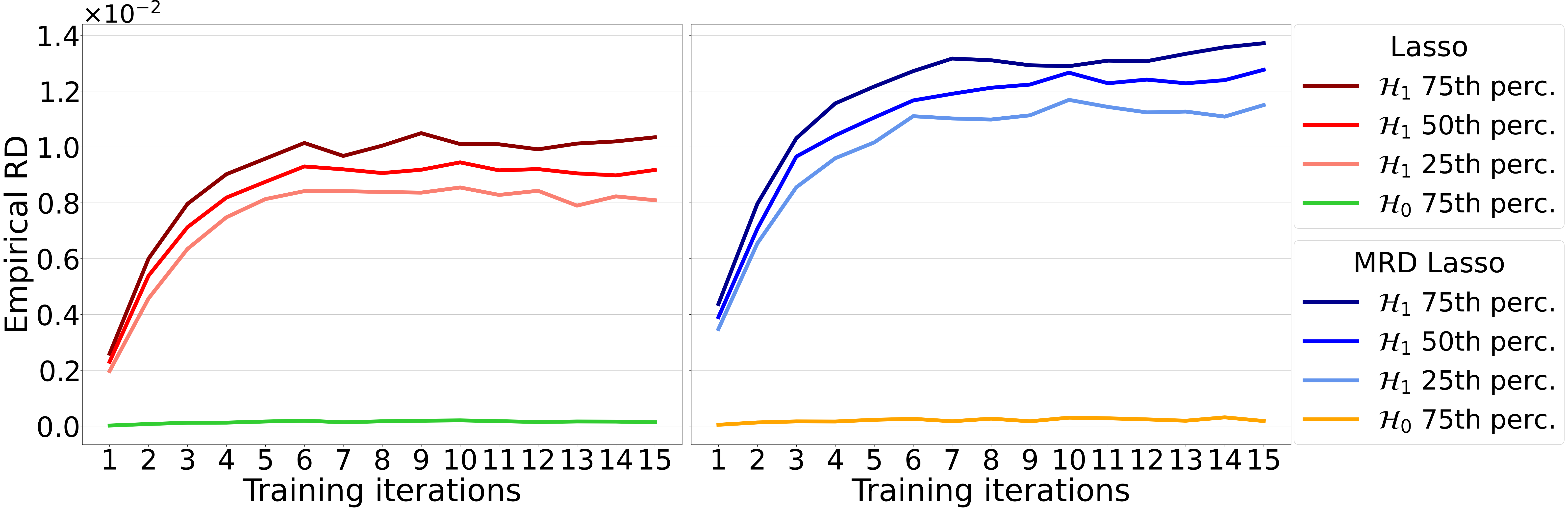}
    \caption{Empirical risk {(MSE)} difference evaluated on test data as a function of the training iterations. Left: baseline lasso model. Right: the proposed MRD method. The red (lasso) and blue (proposed method) curves represent the 25th, 50th, and 75th percentiles of the $\{\widehat{\textrm{RD}}_j : j \in \mathcal{H}_1\}$. The green (lasso) and orange (proposed method) color-shaded curves represent the 75th percentile of $\{\widehat{\textrm{RD}}_j : j \in \mathcal{H}_0\}$. 
    {Each point is evaluated by averaging $\widehat{\textrm{RD}}_j$ over 100  independent data sets.}
    }
  \label{fig:sim_lasso}
\vspace{-0.3cm}
\end{figure}

As a baseline for reference, we fit a lasso regression model on the training set using ADMM \citep{ADMM}, where the lasso penalty parameter is tuned via cross-validation. To analyze the influence of the model on the power of the HRT, we measure the extent to which the \emph{test} mean squared error (MSE) is increased after replacing $X_j$ by its dummy copy $\tilde{X}_j$; this difference serves as an empirical estimate of $\textrm{RD}_j$ \eqref{eq:population_RD}, which we denote by  $\widehat{\textrm{RD}}_j$. Recall that in the context of the HRT, a positive $\widehat{\textrm{RD}}_j$ serves as an evidence against the null, since $\tilde{t}_j > t^*$ in~\eqref{eq:p-value}.
The left panel of Figure~\ref{fig:sim_lasso} presents selected quantiles of $\{\widehat{\textrm{RD}}_j, j\in\mathcal{H}_1\}$ and  $\{\widehat{\textrm{RD}}_j, j\in\mathcal{H}_0\}$ as a function of the ADMM iterations, where the $\widehat{\textrm{RD}}_j$ scores are evaluated on an independent test set. {To reduce randomization, the presented $\widehat{\textrm{RD}}_j$ are averaged over 100 independent train and test sets.} As can be seen, the quantiles that represent the non-null set increase as the training progresses, demonstrating an improvement in the sensitivity of the model to the replacement of these features by their dummy copies. Notice that the lower 25th percentile reaches a final value that is relatively close to the scores that correspond to the null set $\{\widehat{\textrm{RD}}_j : j \in \mathcal{H}_0\}$.
Another important observation is that most of the null scores are approximately equal to zero, implying that the swap of these features with their dummy copies does not lead to significant variations in the test statistics.

Next, we fit the proposed MRD lasso model on the same {100 independent} training data {sets}. The test error obtained by our model is similar to that of the baseline approach; both result in {an averaged} root mean squared error (RMSE) that is equal to 0.94. Also, similarly to the baseline model, the quantiles of the null $\widehat{\textrm{RD}}_j$ scores are close to zero, as shown in the right panel of Figure~\ref{fig:sim_lasso}. This phenomenon exemplifies Corollary~\ref{corollary1}. Turning to the non-null set, we can see that all the quantiles of $\{\widehat{\textrm{RD}}_j : j\in \mathcal{H}_1\}$ obtained by our MRD model increase with the training iterations, reaching final values that are higher than the baseline approach.
This advantage is corroborated by comparing the power of the selection procedure, applied with a target FDR level of $q=0.2$. The empirical power (averaged over 100 independent experiments) obtained by the MRD model is equal to 0.521, higher than the one of the baseline model that equals 0.453. In addition, each method results in an empirical FDR that falls below the nominal rate. Here, the MRD lasso is less conservative, obtaining an FDR of 0.097 whereas the base lasso model results in 0.054.

{We also present the regression coefficients obtained by lasso and MRD lasso in Figure~\ref{fig:bettas} in Supplementary Section~\ref{supp:betas}. In essence, the regression coefficients that correspond to the non-null features obtained by MRD lasso have higher absolute values than those obtained by lasso. By contrast, the regression coefficients that correspond to the null features are closer to zero both for lasso and MRD lasso.}

\subsubsection{Experiments with varying signal strength}
\label{sec:signal_strength}
In the illustrative example presented above we focused on a fixed signal amplitude and sample size, as well as a particular base predictive rule---lasso regression.
We now turn to study the effect of the MRD scheme more systematically, by varying the signal-to-noise ratio and also include additional predictive models. We set $m=400$ as before, but now we increase the auto-correlation parameter to $\rho=0.25$. 

We begin with the polynomial model \textbf{M2} for $Y \mid X$, where we control the signal-to-noise ratio by varying the signal magnitude $c$. The empirical power ($q=0.2$) evaluated for each predictive model is summarized in Table \ref{tab:non_linear_sim_0.2}. As displayed, our MRD approach consistently improves the power of all base models, where the gain is more significant in the low power regime. Notice that the largest improvement is achieved when using lasso as a base model and the smallest gain for the choice of a neural network model (referred as NNet). In terms of the actual power obtained by each method, we typically receive the following relation in this experiment: MRD elastic net $>$ MRD lasso $>$ elastic net $>$ lasso $>$ MRD neural network $>$ neural network. Supplementary Table~\ref{non_linear_sim_0.2_full} presents the and root mean squared error (RMSE) obtained for each machine learning model, showing that the MRD loss tends to achieve RMSE similar to that of the base models. This table also includes comparisons to kernel ridge and random forest, concluding that our MRD elastic net is the most powerful method in this experiment.
Turning to the validity of the selection procedure, following Supplementary Table~\ref{non_linear_sim_0.2_full}, we observe that all methods obtain FDR below the nominal 20\% level, across all data sets. Here, the MRD versions of lasso and elastic net are less conservative in the sense that they obtain higher empirical FDR compared to their base models, whereas the MRD neural network model has comparable FDR to that of its base model. {An in-depth analysis of the above results is presented in Supplementary Section~\ref{sec:pow_q}. There, we present the empirical power and FDR as a function of the FDR target level $q$, as well as Q-Q plots which compare the quantiles of the produced p-values to those of a uniform distribution $U[0,1]$.}

We also perform a similar analysis to the one presented above for the sum of sines model \textbf{M3}  as well as for the linear model \textbf{M1}. We summarize the results in Supplementary Section~\ref{sec:pow_q} for the linear setting and in Supplementary Section~\ref{supp:link_func_} for the non-linear one. The overall trend in both settings is similar to the one presented above: the MRD approach improves the power of the base models, and the FDR is controlled empirically. Lastly, Supplementary Section~\ref{sec:single_hyp} demonstrates that the MRD approach outperforms the baseline methods when testing for a single non-null hypothesis for a wide range of signal strength values. This experiment isolates the effect of the MRD penalty on the resulting power compared to the FDR experiments that account for multiplicity by design. 

\begin{table}[t]
    \caption{Synthetic non-linear data with varying signal strength $c$. The table presents the empirical Power (using FDR target level $q=0.2$) evaluated by averaging over 100 independent experiments, and the relative improvement (\% imp.) of the MRD models. See Supplementary Table~\ref{non_linear_sim_0.2_full} for FDR values.}
     \label{tab:non_linear_sim_0.2}
  \centering
  \resizebox{0.75\textwidth}{!}{\begin{tabular}{@{}cllcllcllc@{}}
    \toprule
    &\multicolumn{3}{c}{Lasso}&
    \multicolumn{3}{c}{Elastic Net}&
    \multicolumn{3}{c}{NNet}
    \\
    \cmidrule(r){2-4}
    \cmidrule(r){5-7}
    \cmidrule(r){8-10}

   $c$ &Base     & MRD     & \% imp. &Base     & MRD     & \% imp.&Base     & MRD     & \% imp.\\
    \midrule
    0.13&0.155&0.243&56.8&0.199&0.269&35.2&0.142&0.156&9.9\\ 
    0.14&0.343&0.435&26.8&0.389&0.454&16.7&0.272&0.291&7.0\\ 
    0.15&0.540&0.606&12.2&0.583&0.621&6.5&0.441&0.455&3.2\\ 
    0.16&0.706&0.740&4.8&0.728&0.753&3.4&0.566&0.606&7.1\\ 
    0.17&0.818&0.841&2.8&0.832&0.848&1.9&0.673&0.692&2.8\\ 
    0.18&0.885&0.898&1.5&0.893&0.900&0.8&0.741&0.768&3.6\\ 
    \bottomrule
  \end{tabular}}
\end{table}

\subsubsection{Experiments with varying sample size}
\label{sec:sample_size}

Thus far we only considered data with a fixed number of samples. In what follows, we study the performance of the proposed MRD approach as a function of the sample size $n$. One can view this as a different way to modify the signal-to-noise ratio in the data, as the power is expected to increase as the sample size increases. We focus on the polynomial model \textbf{M2} for $Y \mid X$ with a correlated design in which the auto-correlation parameter $\rho=0.25$. We generated data with a fixed signal strength $c=0.1$, so that the power will be relatively low for the smallest sample size $n$. Table~\ref{tab:big_n_over_p} summarizes the results, where we deployed only lasso and MRD lasso to ease with the computational load. Following that table we can see that the MRD approach consistently improves the power of the baseline model, while keeping the FDR under control.

\begin{table}[t]
    \caption{Synthetic experiments with non-linear data (\textbf{M2}) with varying number of samples $n$. The empirical FDR (nominal level $q=0.2$) and power are evaluated by averaging over 50 independent experiments. All standard errors are below 0.02.}
     \label{tab:big_n_over_p}
\fontsize{8}{8}\selectfont
  \centering
  \ra{1}

  \begin{tabular}{cllllc}
    \toprule
    &\multicolumn{2}{c}{MRD Lasso}&
    \multicolumn{2}{c}{Lasso}&\multicolumn{1}{c}{\% imp. of}
    \\
    
    \cmidrule(r){2-3}
    \cmidrule(r){4-5}

    $n$ &Power     & FDR     & Power  & FDR& power  \\
    \midrule
    1000 & 0.189 & 0.069 & 0.097 & 0.018&94.5 \\
    1500 & 0.403 & 0.109 & 0.288 & 0.048&39.9 \\
    2000 & 0.648 & 0.106 & 0.563 & 0.046&15.1 \\
    2500 & 0.763 & 0.086 & 0.697 & 0.040&9.5 \\
    3000 & 0.823 & 0.091 & 0.791 & 0.057&4.0 \\
    4000 & 0.927 & 0.116 & 0.921 & 0.061& 0.7\\
    \bottomrule
  \end{tabular}
\end{table}

\subsubsection{Experiments with interaction model}
\label{sec:inter_model}
\begin{figure}[t]
    \centering
    \includegraphics[width=0.7\textwidth]{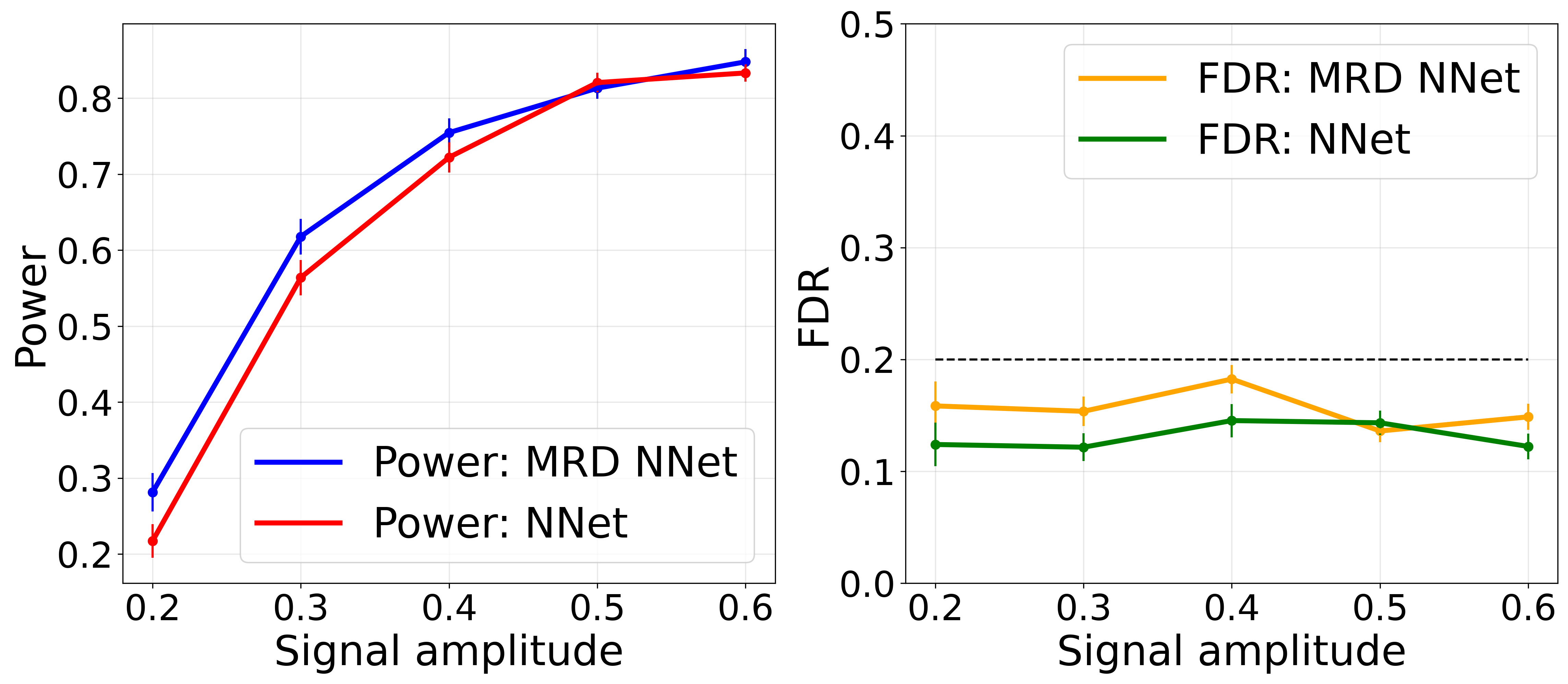}
    \caption{{Synthetic experiments on a sparse interaction model (\textbf{M4}) with varying signal amplitude. Power and FDR ($q=0.2$) are evaluated on $50$ independent trials.}
    } 
    \label{fig:inters_NN}
\end{figure}

In the experiments presented thus far, lasso and elastic net were more powerful than neural networks, even when applied to nonlinear data. In fact, this phenomenon is in line with the experiments reported in \cite[Section~4]{HRT}. We now conduct experiments with a sparse interaction model for the data (\textbf{M4}), for which linear predictive models result in zero power in contrast to neural networks. The results are  summarized in  Figure~\ref{fig:inters_NN}; we do not include linear predictive models as they did not make any discovery.\footnote{Refer to Supplementary Section~\ref{supp:interaction} for additional details on the data generation process and the training strategy.} Notice that the MRD version of the neural network model achieves higher power than the base one, while controlling the FDR.

\subsubsection{Experiments with cross-validation HRT}
\label{sec:cv_hrt}
A limitation of the HRT is the reliance on data splitting, which can lead to a power loss especially when the number of samples is limited. 
When data is scarce, it is more sensible to deploy the cross-validation version of the HRT (CV-HRT) proposed by \cite[][Section 3.1]{HRT}. The CV-HRT leverages the whole data for training and testing, resulting in improved power compared to a single split. 

In what follows, we study the effect of the MRD approach on the CV-HRT in a small sample size regime, where $n<d$. We explore both linear (\textbf{M1}) and nonlinear (\textbf{M2}) data generating functions, where in we set $c=1.5$ and the auto-correlation parameter $\rho=0.25$. 
We implemented the CV-HRT with $K=8$ folds and use only lasso as the base predictive model to ease the computational load. The results are summarized in Table~\ref{tab:n_over_p}, showing that (i) the power is increasing with $n$; (ii) the MRD lasso consistently outperforms the baseline method; and (iii) the FDR is controlled ($q=0.2$).

{In Supplementary Section~\ref{sec:dcrt_compare} we also compare our method to the model-X knockoffs \cite{Knockoffs} and to the state-of-the-art dCRT \cite{dCRT} framework. 
The model-X knockoffs is a multiple testing framework for conditional independence which provides a finite sample FDR control under an arbitrary structure of $(X,Y)$. 
The dCRT, a test for conditional independence, is a clever technique to reduce the computational cost of the original CRT while avoiding data splitting. In a nutshell, this is done by distilling the high-dimensional information that $X_{-j}$ contains on $Y$ into a low-dimensional representation before applying the CRT. Yet, the complexity of dCRT is often higher than that of the CV-HRT as the former requires to fit $d$ different predictive models---one for each leave-one-covariate-out version of the data---compared to $K$ leave-fold-out models in the CV-HRT. 
Section~\ref{sec:dcrt_compare} shows that once combining CV-HRT with our MRD approach we achieve competitive power to that of the dCRT and higher power than the model-X knockoffs.}

\begin{table}[t]
    \caption{Synthetic experiments with linear 
    (\textbf{M1}) and non-linear (\textbf{M2}) data with $n < d$. The empirical FDR (nominal level $q=0.2$) and power are evaluated by averaging over 50 independent experiments, using CV-HRT with 8 folds. All standard errors are below 0.03.}
     \label{tab:n_over_p}
\fontsize{8}{8}\selectfont
  \centering
  \ra{1}

  \begin{tabular}{cllllc}
    \toprule
    \multicolumn{6}{c}{Linear}\\
    \cmidrule(r){1-6}
    &\multicolumn{2}{c}{MRD Lasso}&
    \multicolumn{2}{c}{Lasso}&\multicolumn{1}{c}{\% imp. of}
    \\
    
    \cmidrule(r){2-3}
    \cmidrule(r){4-5}

    $n$ &Power     & FDR     & Power  & FDR& power  \\
    \midrule
    40 & 0.070 & 0.044 & 0.047 & 0.060&48.9 \\
    60 & 0.185 & 0.103 & 0.154 & 0.096&20.1 \\
    70 & 0.367 & 0.105 & 0.304 & 0.106&20.7 \\
    80 & 0.582 & 0.101 & 0.566 & 0.096&2.8 \\
    90 & 0.868 & 0.077 & 0.867 & 0.072& 0.1\\
    \bottomrule
  \end{tabular}
  \quad
  \begin{tabular}{cllllc}
    \toprule
    \multicolumn{6}{c}{Non-Linear}\\
    \cmidrule(r){1-6}
    &\multicolumn{2}{c}{MRD Lasso}&
    \multicolumn{2}{c}{Lasso}&\multicolumn{1}{c}{\% imp. of}
    \\
    
    \cmidrule(r){2-3}
    \cmidrule(r){4-5}

    $n$ &Power     & FDR     & Power  & FDR & power  \\
    \midrule
    80 & 0.067 & 0.081 & 0.055 & 0.056& 21.8\\
    100 & 0.125 & 0.109 & 0.106 & 0.099& 17.9\\
    150 & 0.389 & 0.113 & 0.345 & 0.103& 12.8\\
    200 & 0.648 & 0.080 & 0.616 & 0.073& 5.2\\
    250 & 0.788 & 0.071 & 0.776 & 0.064& 1.5\\
    \bottomrule
  \end{tabular}
\end{table}

\subsubsection{Experiments under model misspecification}
\label{sec:model_misspec}

In the experiments presented thus far we sample the dummy features $\tilde{X}_j$ from the true distribution of $X_j \mid X_{-j}$. In practice, however, we often have only access to an approximation of the unknown conditionals $X_j \mid X_{-j}$. In Supplementary Section~\ref{sec:misspec_lin}, we include experiments under such model misspecification.
There, we fit our MRD models and apply the HRT by sampling approximated dummy features from a data-driven estimation of $P_X$, rather than from the true one. We commence by conducting an experiment in which we control the accuracy of the estimated conditionals. This experiment  demonstrates that an inflation of the empirical FDR occurs only when the sampled dummies are of poor quality, as measured by the covariance goodness-of-fit diagnostic proposed in \cite{DeepKnockoffs}. Next, we study three different settings of increasing difficulty. In the first, $X$ is generated from a correlated multivariate Gaussian, and we sample the approximated dummy features from a multivariate Gaussian whose parameters are estimated from the data. In the second, $X$ is sampled from a Gaussian Mixture Model (GMM), whereas the dummies are sampled from a fitted multivariate Gaussian, as described above. In the third, we sample $X$ from a correlated multivariate Student $t$-distribution. When using a naive multivariate Gaussian fit for $P_X$, we report a violation in FDR control since the estimated conditionals are of low quality.
However, when applying more flexible density estimation method---a Gaussian Mixture Model \cite{GMM}---our method is shown to control the FDR in practice, supporting the discussion from Section~\ref{sec:unknown_p}.

\subsubsection{The effect of the MRD penalty parameter}
\label{sec:additional}
In the interest of space, we refer the reader to Supplementary Section~\ref{supp:lambda} that illustrates the power of the test as a function of the MRD parameter $\lambda$ in low, medium, and high power regimes. This section shows that our automatic approach for selecting $\lambda$ achieves power that is relatively close to an oracle. This section also shows that, in the high power regime ($\approx 0.9$), inappropriate choice of $\lambda$ can reduce power compared to the base model. By contrast, in the low ($\approx 0.5$) and medium ($\approx 0.7$) power regimes, the power of the MRD approach is higher than the power of the base model for all values of $0<\lambda \leq 1$.

\subsection{Real-world application}
\label{sec:real-data}
We now apply our methods to a real-world application that already appears in the knockoff literature \cite{Knockoffs_original,DeepKnockoffs,DeepPink,shaer2022model}. Here, the task is to reliably detect genetic mutations associated with changes in drug resistance among human immunodeficiency viruses (HIV) of type I \cite{HIV};  specifically, we study the resistance to the Lopinavir protease inhibitor drug. The response variable $Y$ represents the log-fold increase in drug resistance measured in the $i$th virus, and each of the features ${X}_j \in \{0,1\}$ indicates the presence or absence of a specific mutation. After applying the pre-processing steps described in \cite{DeepKnockoffs}, the data contains $n=1555$ samples with $d=150$ features. Supplementary Section~\ref{supp:real-data} provides additional details about this data.

Importantly, in contrast to the fully controlled setup from Section~\ref{sec:syn_exp}, here $P_X$ is unknown and thus must be estimated. (Recall that in Supplemetary Section~\ref{sec:misspec_lin} we also study the robustness of MRD to model misspecification.) Here, we approximate $P_X$ by fitting a multivariate Gaussian on all samples $\{X^i\}_{i=1}^n$. While this estimation is most likely inaccurate due to the fact that the features are binary, we follow \cite{DeepKnockoffs} and show via semi-synthetic experiments (for which ${Y \mid X}$ is known) that the selection procedure empirically controls the FDR when sampling $\tilde{X}_j$ from the estimated distribution. Only then, we apply the selection procedure on the real data.

\begin{figure}[t]
    \centering
    \includegraphics[width=0.85\textwidth]{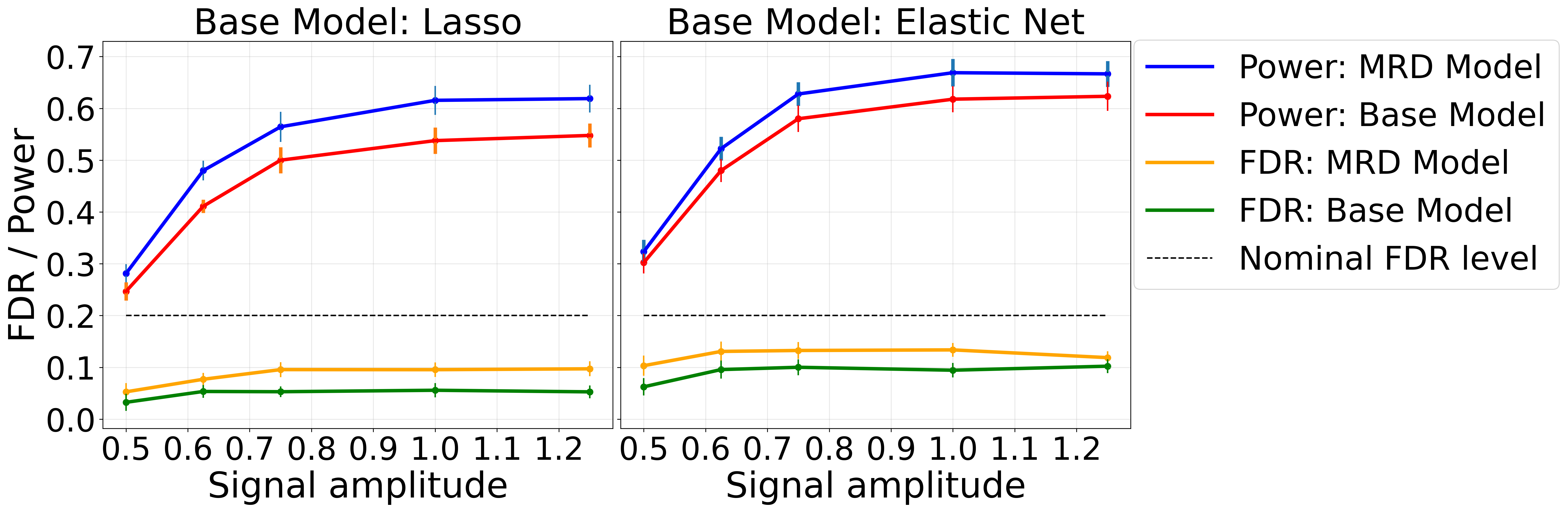}
    \caption{Semi-synthetic experiments with real HIV mutation features and a simulated non-linear response using the generating function  \textbf{M2}. Empirical power and FDR with $q=0.2$ are evaluated over $20$ random train/test splits. 
    } 
    \label{fig:non_lin_semi}
\vspace{-0.15cm}
\end{figure}

\paragraph{Experiments with semi-synthetic data}
We generate a semi-synthetic data set by simulating $Y$ as described in~\eqref{eq:y-mid-x} while treating the real $X$ as fixed, using the polynomial non-linear generating function \textbf{M2}. Next, we randomly split the data into training and testing sets of equal size, and repeat the same experimental protocol, data normalization, and training strategy  used in the synthetic experiments from Section~\ref{sec:syn_exp}.
(For the neural network base model, we slightly modified the number of training epochs due to overfitting; further details are in Supplementary Section~\ref{supp:models}.) Figure~\ref{fig:non_lin_semi} displays the FDR and power obtained by lasso and elastic net as a function of the signal amplitude; a similar figure that describes the results obtained by the base and MRD neural network models is presented in Supplementary Section~\ref{supp:semi_non_linear}. Following Figure~\ref{fig:non_lin_semi}, we can see that all methods control the FDR ($q=0.2$), demonstrating the robustness of the selection procedure to model misspecification. Regarding efficiency---our proposal consistently improves the power of the base models. In contrast to the fully synthetic experiments (Table~\ref{tab:non_linear_sim_0.2}), here the gap between the power of the base model and its MRD version increases as a function of the signal amplitude. Additional semi-synthetic experiments with a linear model that also explore the effect of the sample size are presented in Supplementary Sections \ref{supp:semi_linear} and \ref{supp:semi_vary_n}. Both demonstrate that a similar gain in performance is obtained by our MRD method.


 \begin{figure}
  \begin{center}
    \includegraphics[width=0.5\textwidth]{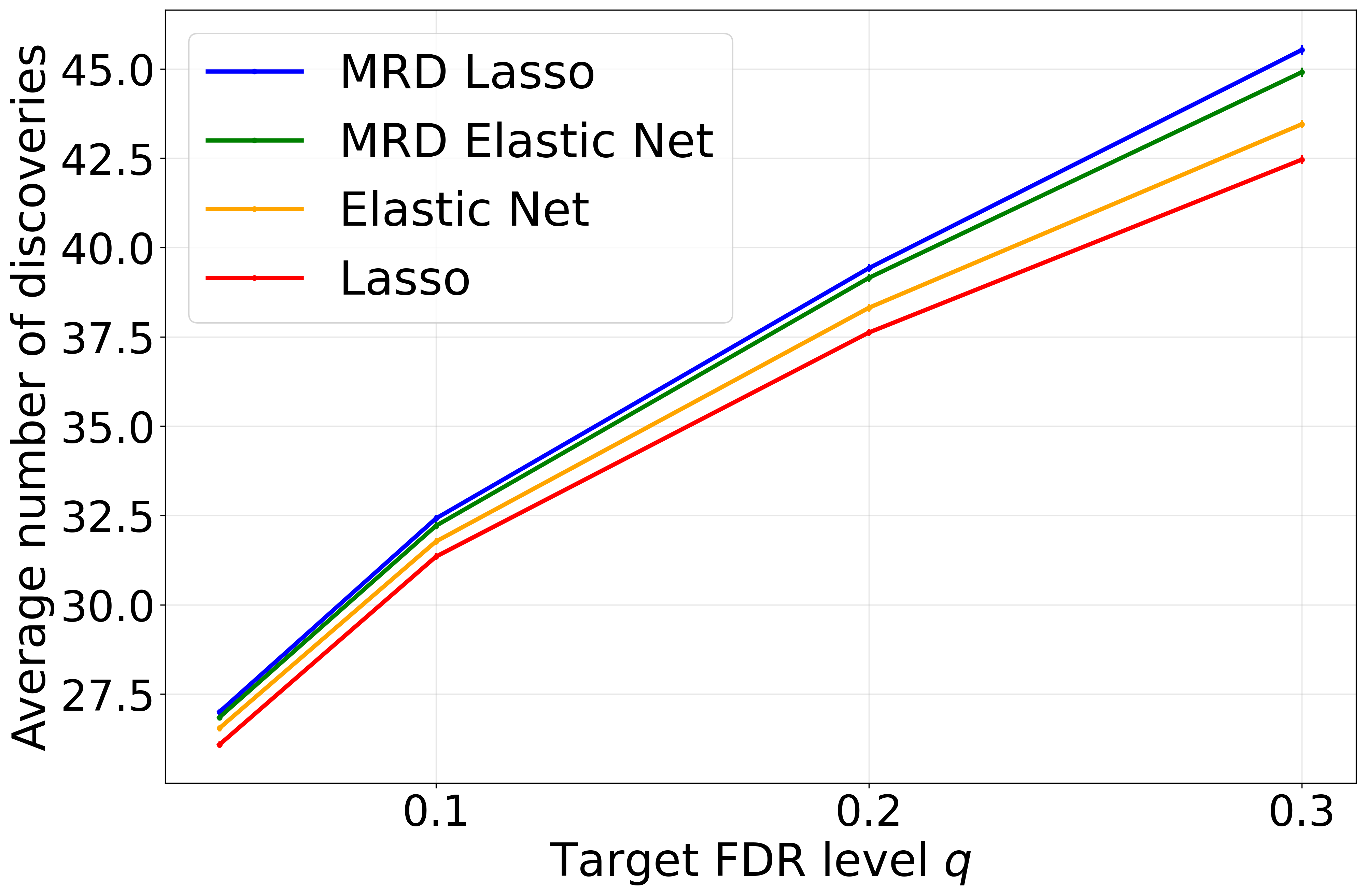}
  \end{center}
  \caption{The average number of drug-resistance mutations in the HIV discovered by base/MRD models.
}
  \vspace{-0.1cm}
  \label{fig:real}
\end{figure}

\paragraph{Experiments with full real data} Next, we perform variable selection on the real data, i.e., with the real $(X,Y)$ pairs. 
Here, we only use lasso and elastic net as base models, and deploy the same training strategy and data-normalization step described in the semi-synthetic experiments. We repeat the variable selection pipeline for 1000 independent train/test splits and present the results in Figure~\ref{fig:real}. 
As portrayed, our MRD models lead to more findings than the baseline methods for various FDR levels, which aligns with the semi-synthetic experiment results. Here, this gap increases with the target FDR level.



\section{Discussion} \label{sec:conclusion}

This paper presents a novel training framework to increase the power of the model-X randomization test. Our method has an advantage over the common practice for which the model-fitting step is not tailored to maximize the power of the HRT. Focusing on FDR control, through experiments, we demonstrate that our method consistently outperforms a wide range of base predictive models in various settings.

{This paper advocates conditional independence testing as a statistical approach to avoid the selection of ``spurious features'', i.e., null features that are only correlated with the non-null ones. Yet, conditional independence testing has an inherent limitation: it is fundamentally impossible to distinguish nearly identical features features. Put differently, it inevitable that we would suffer from low power when analyzing data with extremely correlated features. Figure~\ref{fig:varying_rho} illustrates this phenomenon, where we apply the HRT for multivariate Gaussian $X$ with increasing correlation among the features, controlled by the auto-correlation coefficient $\rho$; see Supplementary Section~\ref{sec:varying_rho} for more details. As portrayed, the power of the test is reduced with the increase of $\rho$, making the selection procedure fundamentally harder. Observe that MRD lasso and MRD elastic net yield higher power than their vanilla counterparts, where the gap between them is reduced as the problem becomes fundamentally harder.}

{One way to deal with the hardness of this problem it to group (or even prune) features that are nearly identical in a given data. For example, \cite{sesia2019gene} apply hierarchical clustering to identify groups of features in such a way that the correlation between the features across the two clusters is not too high. Then, one can apply the selection procedure only on the cluster representatives, which are substantially less correlated by design. This strategy increases power, however at the cost of reducing the resolution of the findings \citep{sesia2019gene}. Rather than testing at the level of individual features, the test is applied to groups of features among which there may be one or more important variables. We believe it will be of great importance to continue this line of work in combination with our MRD approach, and offer practical tools that allow the use of conditional independence testing even for highly correlated data. }

In a broader view of the controlled variable selection problem, the model-X knockoff filter \citep{Knockoffs} provides a rigorous FDR control in finite samples under an arbitrary dependency structure of $(X,Y)$. This stands in contrast with the BH procedure that provably controls the FDR under independence of the p-values constructed by the HRT; see discussion in Section~\ref{sec:setup}. With that said, sampling the knockoff dummies is more challenging and the model fit must be done by augmenting the original features and their knockoff dummies, which increases the dimensions of the problem. Therefore, extending our proposal to work with the knockoffs framework is not straightforward (to say the least), but, at the same time, of great interest.

\section*{Acknowledgments}

We sincerely thank Matteo Sesia for insightful comments about an earlier manuscript draft. S.S and Y.R. were supported by the Israel Science Foundation (grant 729/21). Y.R. also thanks the Career Advancement Fellowship, Technion, for providing research support.

\bibliographystyle{unsrt}
\bibliography{MRD}
\medskip

\appendix

\section{Supplementary proofs}
\label{supp:proofs}

\begin{proof}[Proof of Proposition~1] For completeness, we prove Proposition~\ref{prop1} by following the steps from~\citealp[Proposition 1]{odds}. We use discrete random variables for simplicity.
Denote by $x' = x_{-j}$ and observe that
{
\begin{align*}
    \mathbb{P}(Y = y, X_j = x_j, X_{-j} = x') &= \mathbb{P}(Y = y, X_j = x_j \mid X_{-j}= x') \cdot \mathbb{P}(X_{-j} = x') \\
   \text{(Since $Y \indep X_j \mid X_{-j}$)\quad} &= \mathbb{P}(Y = y \mid X_{-j} = x') \cdot \mathbb{P}(X_j = x_j \mid X_{-j} = x') \cdot \mathbb{P}(X_{-j} = x') \\
   \text{(By construction of $\tilde{X}_j$)\quad} &= \mathbb{P}(Y = y \mid X_{-j} = x') \cdot \mathbb{P}(\tilde{X}_j = x_j \mid X_{-j} = x') \cdot \mathbb{P}(X_{-j} = x') \\
    \text{(Since $Y \indep \tilde{X}_j \mid X_{-j}$)\quad} &= \mathbb{P}(Y = y, \tilde{X}_j = x_j \mid X_{-j} = x') \cdot \mathbb{P}(X_{-j} = x') \\
    &= \mathbb{P}(Y = y, \tilde{X}_j = x_j, X_{-j} = x').
\end{align*}
}
\end{proof}

\begin{proof}[Proof of Proposition~2]

Following Corollary~\ref{corollary1}, for any null feature $\beta_j = 0$, we know that the following holds
\begin{align}
    & \mathbb{E}[T(Y,\tilde{X}_j,X_{-j}; \hat{\beta}) - T(Y,X_j,X_{-j}; \hat{\beta})] = 0,
\end{align}
for any value of $\hat{\beta}_j$. We now turn to study the effect of $\hat{\beta}_j, j\in\mathcal{H}_0$ on a non-null feature $k \in \mathcal{H}_1$, by re-writing the test statistic:
\begin{align}
    \label{eq:original_t_prop2_proof}
    \mathbb{E}[T(Y,{X}_k,X_{-k}; \hat{\beta})] & = \mathbb{E}[(Y - {X}_j \hat{\beta}_j - {X}_k \hat{\beta}_k -  {X}_{-(j,k)}^T\hat{\beta}_{-(j,k)})^2] \\
    & = \mathbb{E}[({X}_j {\beta}_j + {X}_k {\beta}_k + {X}_{-(j,k)}^T{\beta}_{-(j,k)} + \\
    & \quad\quad\quad \epsilon - {X}_j \hat{\beta}_j - {X}_k \hat{\beta}_k - {X}_{-(j,k)}^T\hat{\beta}_{-(j,k)})^2] \\
    & = \mathbb{E}[({X}_k ({\beta}_k - \hat{\beta}_k) + {X}_{-(j,k)}^T({\beta}_{-(j,k)} - \hat{\beta}_{-(j,k)}) + \epsilon - {X}_j \hat{\beta}_j)^2] \\
    & = \mathbb{E}[({X}_k ({\beta}_k - \hat{\beta}_k))^2] + \mathbb{E}[({X}_{-(j,k)}^T({\beta}_{-(j,k)} - \hat{\beta}_{-(j,k)}))^2] + 
    \\ & \quad \  \mathbb{E}[\epsilon^2] + \hat{\beta}_j^2\mathbb{E}[{X}_j^2].
\end{align}
The third equality uses the fact that $\beta_j=0$ and the last one relies on the assumptions that $\mathbb{E}[X_j,X_k] = 0 $ for all $1 \leq j \neq k \leq d$ and $\mathbb{E}[X_j,\epsilon] = 0 $ for all $1 \leq j \leq d$. Similarly, since $\mathbb{E}[{X}_j \tilde{X}_k] = 0$ we have also that
\begin{align}
    \mathbb{E}[T(Y,\tilde{X}_k,X_{-k}; \hat{\beta})] &= \mathbb{E}[({X}_k{\beta}_k - \tilde{X}_k\hat{\beta}_k))^2] + \mathbb{E}[({X}_{-(j,k)}^T({\beta}_{-(j,k)} - \hat{\beta}_{-(j,k)}))^2] + 
    \\ & \quad \  \mathbb{E}[\epsilon^2] + \hat{\beta}_j^2\mathbb{E}[{X}_j^2].
\end{align}
As a result, we get that $\hat{\beta}_j, j\in\mathcal{H}_0$ does not influence the risk discrepancy of the null and non null features.
To complete the proof, we now show that any choice of $ \hat{\beta}_j \neq 0$ can only increase the MSE term in the objective function. Applying similar steps as in \eqref{eq:original_t_prop2_proof}, we get
\begin{align}
    \mathbb{E}[(Y - X^T\hat{\beta})^2] = \mathbb{E}[({X}_{-j}^T({\beta}_{-j} - \hat{\beta}_{-j}))^2] + \mathbb{E}[\epsilon^2] + \hat{\beta}_j^2\mathbb{E}[{X}_j^2],
\end{align}
which is minimized for the choice of $\hat{\beta}_j = 0$.
\end{proof}

\section{Supplementary analysis of the optimization algorithm}
\label{supp:optim}
In this section, we follow \cite{DeepKnockoffs,sanjabi2018solving} and provide a weak form of convergence to the general learning scheme presented in Algorithm \ref{alg:general_optimization} of the main manuscript.
Denote by $f_{\theta_k}$ a predictive model, parameterized by $\theta_k$ (e.g., the weights and biases of a neural network), at step $k$ of the optimization procedure. Recall that in Algorithm~\ref{alg:general_optimization}, in each iteration we choose a random subset of indices $\mathcal{P}_k$ and generate a \emph{fresh} dummy copy $\tilde{X}_{j}^i$ of ${X}_{j}^i$ for all $j \in \mathcal{P}_k$. For ease of notation, we denote by $\tilde{\textbf{X}}_k \in \mathbb{R}^{m \times |\mathcal{P}|}$ the dummy matrix with the new $\tilde{X}_{j}^i$ as its $(i,j)$'s entry. Thus, conditional on $(\mathcal{P}_k,\tilde{\textbf{X}}_k)$ the objective function 
$$\mathcal{J}(\textbf{X},\textbf{Y},f_{\theta_k},\mathcal{P}_k,\tilde{\textbf{X}}_k) =(1-\lambda) \mathcal{J}_{\textrm{base}}(\textbf{Y},\textbf{X}; f_{\theta_k}) + \frac{\lambda}{|\mathcal{P}_k|} \sum_{j \in \mathcal{P}_k} {\mathcal{D}( z(f_{\theta_k}), \tilde{z}_j(f_{\theta_k}))},$$
is a deterministic function of $\theta_k$. Now, we can define 
\begin{equation}
    \label{eq:optim1}
    \mathcal{J}_{\theta_k}=\mathbb{E}_{\mathcal{P}_k,\tilde{\textbf{X}}_k}[{\mathcal{J}(\textbf{X},\textbf{Y},f_{\theta_k},\mathcal{P}_k,\tilde{\textbf{X}}_k) \mid \theta_k}]
\end{equation}
and denote by $\nabla\mathcal{J}_{\theta_k}$ its gradient with respect to $\theta_k$. As depicted in Algorithm~\ref{alg:general_optimization}, we estimate $\nabla\mathcal{J}_{\theta_k}$ by sampling one realization of $\mathcal{P}_k$ and $\tilde{\textbf{X}}_k$: 
\begin{equation}
\label{eq:g_t}
    g_t = \nabla\mathcal{J}(\textbf{X},\textbf{Y},f_{\theta_k},\mathcal{P}_k,\tilde{\textbf{X}}_k),
\end{equation}

resulting in an unbiased estimate of $\nabla\mathcal{J}_{\theta_k}$. Since all random variables are observed at the $k$th step, $g_t$ from \eqref{eq:g_t} is deterministic and backpropogation can be applied to update $\theta_{k+1} \leftarrow \theta_{k} - \eta g_t$, where $\eta$ is the learning rate (step size). Our analysis assumes that there exists a finite Lipschitz constant $L$, which satisfies the following relation for all $\theta',\theta''$:

$$
||\nabla\mathbb{E}_{\mathcal{P},\tilde{\textbf{X}}}[{\mathcal{J}(\textbf{X},\textbf{Y},f_{\theta},\mathcal{P},\tilde{\textbf{X}}) \mid \theta']} -\nabla\mathbb{E}_{\mathcal{P},\tilde{\textbf{X}}}[{\mathcal{J}(\textbf{X},\textbf{Y},f_{\theta},\mathcal{P},\tilde{\textbf{X}}) \mid \theta'']}||_2 < L||\theta'-\theta''||_2.
$$

Denote $\Delta = \frac{2}{L} \sup (\mathcal{J}_{\theta_1} -\mathcal{J}^*)$, where $\mathcal{J}_{\theta_1}$ is the objective obtained by the initialized model, i.e., by assigning $\theta_1$ in \eqref{eq:optim1}, and $\mathcal{J}^*$ is a lower bound uniformly the value of  $\mathcal{J}_{\theta_k}$, for all $k$. The supremum is taken over all possible values of $\mathcal{P},\tilde{\textbf{X}}$ and $\theta_1$. Suppose we can uniformly bound from below the RD term, which holds for our choice of $\mathcal{D}$ from Section \ref{sec:gen_scheme}, since $\mathcal{D}(z,\tilde{z}_j) = \sigma(z-\tilde{z}_j) > 0$. Suppose further that we can uniformly lower bound the objective of the base predictive model; for example, the MSE loss is bounded from below by 0. With these notations in place, we can immediately invoke Theorem~2 from \cite{DeepKnockoffs}, which is stated below formally for convenience. In short, Theorem~\ref{theo:optim1} provides us a weak form of convergence as it shows that the average squared norm of the gradient $\mathbb{E}[||\nabla \mathcal{J}_{\theta_k} ||_2^2 \mid \theta_1]$ decreases as $\mathcal{O}(1/\sqrt{K})$; a detailed discussion on this result is provided in \cite{DeepKnockoffs}; see also \cite{sanjabi2018solving}.
\begin{theorem}{(Theorem 2, \cite{DeepKnockoffs})}
\label{theo:optim1}
Consider a fixed training set $\textbf{X},\textbf{Y}$, and use the notation from above. Assume
$$
\mathbb{E}[||g_t - \nabla\mathcal{J}_{\theta_k}||_2^2 \mid \theta_k] \leq \sigma^2, \quad \forall k \leq K,
$$
for some $\sigma \in \mathbb{R}$. Then for any initial $\theta_1$ and a suitable value of $\Delta$,
$$
\frac{1}{K} \sum_{k=1}^K {\mathbb{E}[||\nabla \mathcal{J}_{\theta_k} ||_2^2 \mid \theta_1]} \leq \frac{1}{K} \frac{L\Delta}{\eta(2-L\eta)}+\frac{L\sigma^2\eta}{(2-L\eta)}.
$$
Specifically, choosing $\eta=\min \{\frac{1}{L},\frac{\eta_0}{\sigma\sqrt{K}}\}$ for some $\eta_0 > 0$ gives

$$
\frac{1}{K} \sum_{k=1}^K {\mathbb{E}[||\nabla \mathcal{J}_{\theta_k} ||_2^2 \mid \theta_1]} \leq \frac{L^2\Delta}{K} + \bigg(\eta_0 +\frac{\Delta}{\eta_0} \bigg) \frac{L\sigma}{\sqrt{K}}.
$$

\end{theorem}

\section{Supplementary details on fitting sparse linear regression models}
\label{supp:sparse}
Algorithm~\ref{alg:admm} presents our specialized learning scheme for linear regression model. This algorithm minimizes the following objective:
\begin{align}
\hat{\beta} = &\underset{\beta, \textrm{v}}{\textup{argmin}} \  (1-\lambda)\bigg(\frac{1}{2m} || \textbf{X}\mathrm{v} - \textbf{Y}||^2_2 + \alpha_1 ||\beta||_1 + \frac{\alpha_2}{2}||\beta||_2^2\bigg) + \frac{\lambda}{d} \sum_{j=1}^{d} {\mathcal{D}( z(\mathrm{v}), \tilde{z}_j(\mathrm{v}))} \\ &\textrm{subject to} \quad \beta=\mathrm{v},
\label{eq:data_splitting}
\end{align}
using ADMM \cite{ADMM}.

\begin{algorithm}[h]
\textbf{Input}: Training data $\{(X^i,Y^i)\}_{i=1}^m$; elastic net hyperparameters $\alpha_1,\alpha_2$; ADMM hyperparameter $\rho$; MRD hyperparameter $\lambda$;  number of features to generate $1 \leq N \leq d$.

\begin{algorithmic}[1]
\STATE $\textrm{Initialize } \beta^0 \leftarrow \textbf{0}, \mathrm{v}^0 \leftarrow \textbf{0},  \mathrm{u}^0 \leftarrow \textbf{0};$ 
where $\mathrm{u}$ is the scaled Lagrange multipliers vector.

\FOR{$k = 1,\dots,K$}
\STATE $\textrm{Sample a random subset of features } \mathcal{P} \subseteq \{1 \cdots\ d\} \textrm{ of the size } N$.

\STATE $\textrm{Generate a dummy } \tilde{X}_j^i \textrm{for each training point } X_i \textrm{ and for each } j \in \mathcal{P}$.

\STATE $\textrm{Update } \mathrm{v} \textrm{ using gradient descent and back-propagation:}$
$$
\mathrm{v}^{k+1} \leftarrow \underset{\mathrm{v}}{\textup{argmin}} \frac{1-\lambda}{2m}|| \textbf{X}\mathrm{v} - \textbf{Y}||^2_2 +\frac{\lambda}{|\mathcal{P}|} \sum_{j \in \mathcal{P}} {\mathcal{D}( z(\mathrm{v}), \tilde{z}_j(\mathrm{v}))} + \frac{\rho}{2} || \mathrm{v} -\beta^k + \mathrm{u}^k ||_2^2.
$$

\STATE Update $\beta$ using the proximal mapping of elastic net: 
$$\beta^{k+1}\leftarrow \frac{1}{1+\alpha_2}S_{\alpha_1/\rho}(\mathrm{v}^{k+1} + \mathrm{u}^k/\rho),$$

where 
$S_\tau(x) = \max\{0, x-\tau \} - \max\{0, -x-\tau \}.$

\STATE Update $\mathrm{u}$ by setting $\mathrm{u}^{k+1} \leftarrow \mathrm{u}^k + \mathrm{v}^{k+1} - \beta^{k+1}.$

\ENDFOR
\end{algorithmic}
\vspace{0.1cm}
\textbf{Output}: A regression coefficient vector $\hat{\beta}$.
\caption{Fitting MRD elastic net via ADMM}
\label{alg:admm}
\end{algorithm}

\section{Supplementary details on predictive models and training strategy}
\label{supp:models}

Section \ref{sec:experiments} of the main manuscript compares the performance of several regression models on synthetic and real data sets. Below, we provide additional details on each method.
\begin{itemize}
    \item \textbf{Lasso}: we fit a lasso regression model using Python's \texttt{sklearn} package and tuned the lasso penalty parameter via 5-fold cross-validation.
    \item \textbf{Elastic net}: an elastic net model is fitted using the same Python package and training strategy described above.
\item \textbf{Neural network (NNet)}: the network architecture consists of two fully connected layers with a hidden dimension of $16$; the model is fitted using Adam optimizer~\cite{Adam} with a fixed learning rate of $0.005$. To avoid over-fitting, a dropout regularization~\cite{Dropout} is deployed, with a rate of $0.5$. We also use CancelOut~\cite{CancelOut} regularization to promote sparsity in the entry layer.\footnote{The implementation is available online at \url{https://github.com/unnir/CancelOut}} In Section~\ref{sec:syn_exp}, we train the network for $60$ epochs, whereas in Section~\ref{sec:real-data} we reduced this number due to over-fitting. Specifically, we use $15$ epochs for $c\in\{0.5,0.625,0.75\}$, and $10$ epochs for $c\in\{1,1.25\}$, where $c$ is the signal strength of the semi-synthetic data.

\item \textbf{MRD lasso}: we fit the model by minimizing \eqref{mrd_objective} in combination with lasso regularization using Algorithm~\ref{alg:admm}. The lasso penalty parameter is equal to the one used in \textbf{lasso}. The MRD penalty $\lambda$ is set to be equal to $\min\{0.8, 0.8\cdot\texttt{MSE-Validation}\}$, where \texttt{MSE-Validation} is the average MSE evaluated on the 5 folds used to tune the hyperparameter of the baseline lasso model. Regarding the hyperparameter of the ADMM framework, we set $\rho=1$ and the stopping criteria parameters $\epsilon^{\mathrm{rel}} = 10^{-3},\epsilon^{\mathrm{abs}}=5\cdot10^{-4}$; we deployed the stopping rule suggested in \cite[Section 3.3]{ADMM}.

\item \textbf{MRD elastic net}: here, we combine the elastic net penalty with our RD loss and use the same training strategy described in \textbf{MRD lasso}.

\item \textbf{MRD neural network}: we choose the same architecture, optimizer, and hyperparameters outlined in \textbf{neural network}, and fit the model as described in Algorithm~\ref{alg:general_optimization}. We set the MRD penalty $\lambda$ to be equal to $\min\{0.8, 0.8\cdot\texttt{MSE-Validation}\}$, where the \texttt{MSE-Validation} is evaluated as follows. First, we split the training data into training (80\%) and validation (20\%) sets. Then, we fit the base neural network model (without the penalty) on the training data, and use this model to compute the \texttt{MSE-Validation} on the validation set.
\end{itemize}

We run all the numerical experiments on our local cluster, which consists of two AMD servers with 24 CPU cores each.

\section{Supplementary details on the effect of $\lambda$}
\label{supp:lambda}
\begin{figure}[h]
    \centering
    \includegraphics[width=\textwidth]{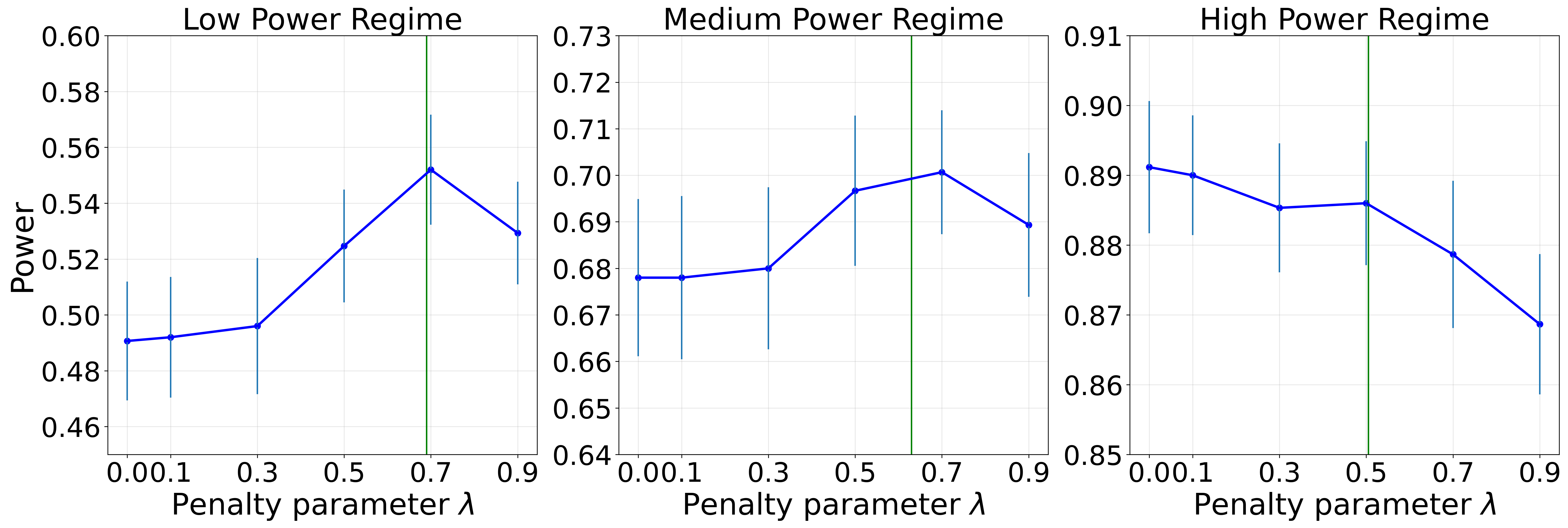}
    \caption{The effect of the MRD regularization strength on power, demonstrated using a synthetic data with correlated multivariate Gaussian features and a simulated response that follows a non-linear model. Each panel displays the empirical power (evaluated by averaging over 50 independent experiments) as a function of the MRD penalty parameter $\lambda$, using lasso as the base predictive model. The green vertical line represents the value obtained by our proposed automatic approach for choosing $\lambda$. Left: low power regime for which the signal amplitude $c=0.14$. Middle: low power regime for which the signal amplitude $c=0.15$. Right: high power regime for which the signal amplitude $c=0.18$.} 
    \label{fig:varying_lambda}
\end{figure}

Section~\ref{supp:models} of the Supplementary Material presents a simple way to choose the penalty parameter $\lambda$. Here, we demonstrate the impact of different values of $\lambda$ on the performance of the MRD approach. Figure~\ref{fig:varying_lambda} presents the empirical power obtained by using the MRD lasso, as a function of the penalty parameter $\lambda$. The vertical green line represents the choice of our approach for choosing $\lambda$. Here, we use the same non-linear synthetic data as in Section~\ref{sec:illustrative_example}, with three different signal amplitude values, $c=0.14$, $c=0.15$ and $c=0.18$, representing low, medium and high power regimes. As can be seen, in the low and medium power regimes, the choice of every presented value of $\lambda$ results in higher power than the base model (i.e., $\lambda=0$). However, in the high power regime, higher values of $\lambda$ reduces the power of the base models. This phenomenon corroborates our approach to choose $\lambda$ to be proportional to the prediction error of the base model, as presented in Supplementary Section~\ref{supp:models}.

Following that figure, we can also see that our proposed automatic approach for choosing $\lambda$, returns a penalty parameter that is fairly close the best one.

\section{The effect of MRD on lasso's regression coefficients}
\label{supp:betas}
\begin{figure}
     \centering
    \begin{subfigure}[b]{0.48\textwidth}
         \centering
         \includegraphics[width=\textwidth]{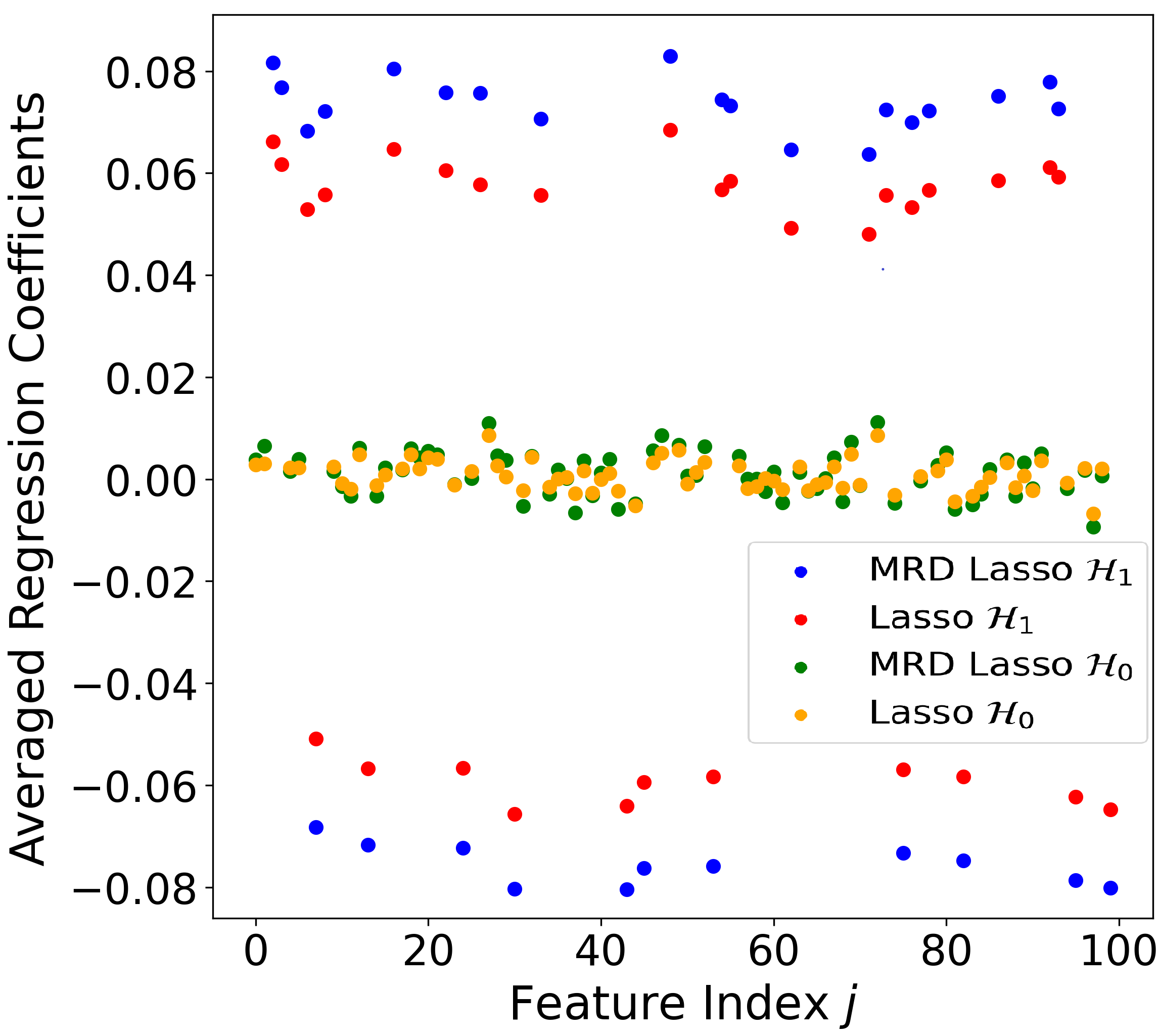}
         \caption{}
         \label{fig:bettas_avg}
     \end{subfigure}
     \hfill
    \begin{subfigure}[b]{0.445\textwidth}
         \centering
         \includegraphics[width=\textwidth]{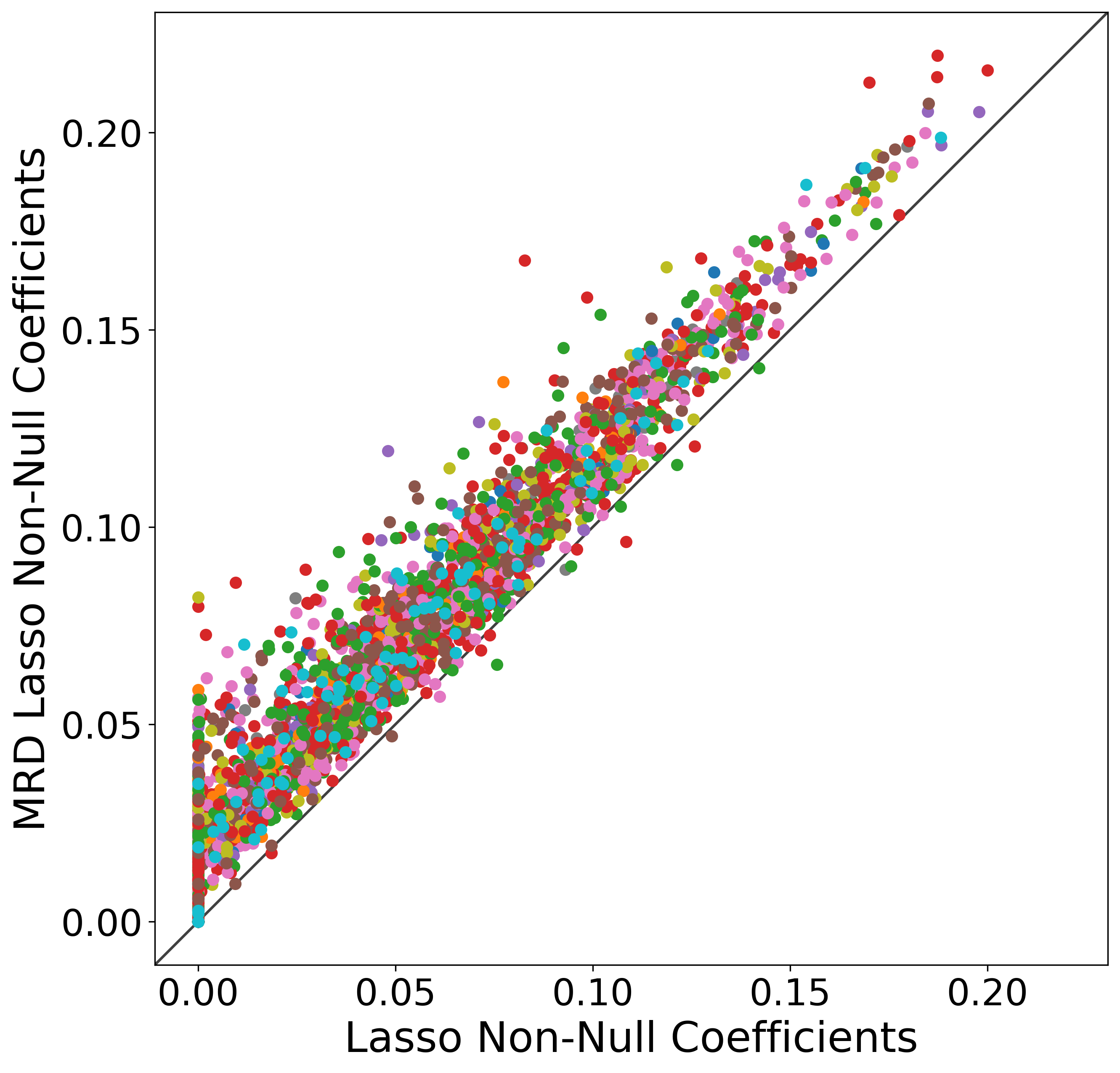}
         \caption{}
         \label{fig:bettas_vs_bettas}
     \end{subfigure}
     
        \caption{{\textbf{Comparison between MRD lasso regression coefficients and vanilla lasso regression coefficients}. The predictive models are fitted on the synthetic non linear data from Section~\ref{sec:illustrative_example}. Left: the regression coefficients of lasso and MRD lasso for all features. Each point in this panel is obtained by averaging the estimated regression coefficient over 100 independent trials. Green and orange dots represent null features; blue and red dots represent non-null features. Right: scatter plot of the magnitude of MRD lasso regression coefficients as a function of the vanilla lasso regression coefficients for non-null features. Each color represent a different non-null feature.}}
        \label{fig:bettas}
\end{figure}

{In Section~\ref{sec:linear_mrd} we describe our learning procedure for the choice of sparse linear regression as the base objective. Here, we demonstrate the impact of the MRD approach on the estimated regression coefficients. Figure~\ref{fig:bettas_avg} presents the average of the estimated regression coefficients of lasso and MRD lasso from the experiment presented in Section~\ref{sec:illustrative_example}.
As can be seen, on average, the estimated regression coefficients of the null features that correspond to the MRD lasso are slightly larger than those of lasso, but both are relatively close to zero.
By contrast, the coefficients of the non-null features have a larger magnitude, where it is evident that the MRD approach further increases the coefficients' magnitude.  Figure~\ref{fig:bettas_vs_bettas} also presents the magnitude of the estimated coefficients of  MRD lasso that correspond to the non-null features as a function of lasso coefficients. Each color is for a different important feature, so there are 100 points per feature, one for a different trial. We can see that the MRD lasso's coefficients are with larger magnitude than those of lasso, where the amplification of the lower coefficients is more significant than of the higher ones. Specifically, it can be observed that in several cases lasso nullifies coefficients of important features, whereas MRD lasso yields non-zero values (the opposite is not true).}

\clearpage
\section{Supplementary details on synthetic experiments}
\label{supp:syn_exp}

\subsection{Further Analysis of The Numerical Experiments}
\label{sec:pow_q}

\paragraph{Linear $Y\mid X$  and additional comparisons} In Table~\ref{tab:non_linear_sim_0.2} of the main manuscript we compared the performance of MRD lasso, MRD elastic net and MRD NNet to their base models by varying the signal strength while fixing the target FDR level to $q=0.2$. For completeness, we provide here the root mean squared error (RMSE) and the empirical FDR obtained by each model as well as additional comparisons to other baseline methods. We also present here the empirical power, FDR and RMSE of our experiments in the linear setting, where $Y \mid X$ follows \textbf{M1}. 
Table~\ref{linear_sim_0.2_full} focuses on the linear setting, where we include a ridge regression model in addition to lasso and elastic net. The penalty parameter of the ridge regression model is tuned via 5-fold cross-validation. Following that table, we can see that the empirical power obtained by the ridge model does not exceed the ones achieved by our proposed MRD methods. Observe also that the RMSE scores obtained by MRD lasso and MRD elastic net are similar to that of the corresponding baseline models. Turning to the non-linear case, Table~\ref{non_linear_sim_0.2_full} summarizes the performance of random forest and kernel ridge regression in addition to lasso, elastic net, and neural network. We fit the kernel ridge regression model using the \texttt{sklearn} software package, where we choose a polynomial kernel of degree 3 and a fixed penalty parameter that equal to 1. The random forest model is also fitted using \texttt{sklearn}, where we choose the software's default hyperparameters except the number of trees and the maximum depth of each tree; we set both to 100. In terms of power, the kernel ridge model outperforms the baseline models, however, it does not exceed the performance of MRD lasso and MRD elastic net. The random forest model is less competitive and yields lower power compared to the other methods, across all experiments. Regarding the RMSE, overall we can see that the MRD models perform similarly to their baseline counterparts, except the case of a neural network base model. There, the RMSE of the MRD neural network is slightly improved.

\begin{table}
  \caption{Synthetic linear data with varying signal strength $c$. The table presents the empirical Power (using FDR target level $q=0.2$) evaluated by averaging over 100 independent experiments, and the relative improvement (\% imp.) of the MRD models. Standard errors are below 0.012.}
  \label{linear_sim_0.2_full}
  \centering
  \resizebox{\textwidth}{!}{
  \begin{tabular}{@{}clllllllllllllll@{}}
    \toprule
    &\multicolumn{3}{c}{MRD Lasso}&
    \multicolumn{3}{c}{Lasso}&
    \multicolumn{3}{c}{MRD Elastic Net}&
    \multicolumn{3}{c}{Elastic Net}&
    \multicolumn{3}{c}{Ridge}
    \\
    \cmidrule(r){2-4}
    \cmidrule(r){5-7}
    \cmidrule(r){8-10}
    \cmidrule(r){11-13}
    \cmidrule(r){14-16}
    $c$ &Power     & FDR & RMSE     & Power  & FDR  & RMSE    & Power  & FDR  & RMSE  & Power  & FDR  & RMSE  & Power  & FDR & RMSE\\
    \midrule
    0.11 & 0.457 & 0.091 & 1.021 & 0.377 & 0.050 & 0.998 & 0.502 & 0.095 & 0.996 & 0.436 & 0.073 & 0.991 & 0.467 & 0.135 & 0.979\\
    0.12 & 0.578 & 0.097 & 0.995 & 0.514 & 0.053 & 0.977 & 0.604 & 0.095 & 0.975 & 0.560 & 0.073 & 0.972 & 0.578 & 0.141 & 0.955\\
    0.13 & 0.679 & 0.091 & 0.964 & 0.628 & 0.057 & 0.952 & 0.688 & 0.102 & 0.950 & 0.656 & 0.072 & 0.947 & 0.665 & 0.139 & 0.931\\
    0.14 & 0.752 & 0.085 & 0.931 & 0.722 & 0.058 & 0.923 & 0.766 & 0.095 & 0.922 & 0.739 & 0.074 & 0.920 & 0.742 & 0.144 & 0.907\\
    0.15 & 0.821 & 0.095 & 0.897 & 0.790 & 0.059 & 0.892 & 0.823 & 0.094 & 0.892 & 0.805 & 0.072 & 0.891 & 0.804 & 0.145 & 0.884\\
    0.16 & 0.867 & 0.083 & 0.865 & 0.855 & 0.061 & 0.862 & 0.871 & 0.098 & 0.863 & 0.858 & 0.072 & 0.862 & 0.850 & 0.145 & 0.860\\
    0.17 & 0.901 & 0.086 & 0.837 & 0.891 & 0.064 & 0.835 & 0.903 & 0.093 & 0.836 & 0.895 & 0.075 & 0.835 & 0.888 & 0.142 & 0.837\\
    0.18 & 0.933 & 0.082 & 0.811 & 0.925 & 0.069 & 0.810 & 0.932 & 0.098 & 0.811 & 0.926 & 0.073 & 0.811 & 0.917 & 0.143 & 0.815\\
    \bottomrule
  \end{tabular}
  }
\end{table}

\begin{table}
  \caption{Complementary results to those presented in Table~\ref{tab:non_linear_sim_0.2}. All standard errors are below 0.02.}
  \label{non_linear_sim_0.2_full}
  \tiny
  \centering
  \resizebox{\textwidth}{!}{
  \begin{tabular}{@{}cllllllllllll@{}}
    \toprule
    &\multicolumn{3}{c}{MRD Lasso}&
    \multicolumn{3}{c}{Lasso}&
    \multicolumn{3}{c}{MRD Elastic Net}&
    \multicolumn{3}{c}{Elastic Net}
    \\
    \cmidrule(r){2-4}
    \cmidrule(r){5-7}
    \cmidrule(r){8-10}
    \cmidrule(r){11-13}

    $c$ &Power     & FDR & RMSE    & Power  & FDR & RMSE    & Power  & FDR & RMSE    & Power  & FDR  & RMSE \\
    \midrule
    0.13 & 0.243 & 0.095 & 0.979 & 0.155 & 0.04 & 0.973 & 0.269 & 0.098 & 0.964 & 0.199 & 0.048 & 0.967\\
    0.14 & 0.435 & 0.093 & 0.949 & 0.343 & 0.039 & 0.946 & 0.454 & 0.1 & 0.938 & 0.389 & 0.062 & 0.941\\
    0.15 & 0.606 & 0.09 & 0.916 & 0.54 & 0.049 & 0.914 & 0.621 & 0.105 & 0.908 & 0.583 & 0.063 & 0.91\\
    0.16 & 0.74 & 0.092 & 0.882 & 0.706 & 0.053 & 0.879 & 0.753 & 0.105 & 0.878 & 0.728 & 0.074 & 0.879\\
    0.17 & 0.841 & 0.093 & 0.851 & 0.818 & 0.057 & 0.849 & 0.848 & 0.098 & 0.849 & 0.832 & 0.075 & 0.849\\
    0.18 & 0.898 & 0.093 & 0.823 & 0.885 & 0.06 & 0.821 & 0.9 & 0.101 & 0.822 & 0.893 & 0.075 & 0.822\\
    
    \toprule
    &\multicolumn{3}{c}{MRD NNet}&
    \multicolumn{3}{c}{NNet}&
    \multicolumn{3}{c}{Kernel Ridge}&
    \multicolumn{3}{c}{Random Forest}
    \\
    \cmidrule(r){2-4}
    \cmidrule(r){5-7}
    \cmidrule(r){8-10}
    \cmidrule(r){11-13}
    
     &Power     & FDR & RMSE    & Power  & FDR & RMSE    & Power  & FDR & RMSE    & Power  & FDR  & RMSE \\
    \midrule
    0.13 & 0.156 & 0.111 & 0.994 & 0.142 & 0.131 & 1.018 & 0.204 & 0.16 & 0.988 & 0.044 & 0.149 & 0.993\\
    0.14 & 0.291 & 0.119 & 0.964 & 0.272 & 0.132 & 0.984 & 0.383 & 0.134 & 0.956 & 0.087 & 0.109 & 0.986\\
    0.15 & 0.455 & 0.124 & 0.933 & 0.441 & 0.14 & 0.948 & 0.551 & 0.139 & 0.923 & 0.141 & 0.14 & 0.979\\
    0.16 & 0.606 & 0.133 & 0.902 & 0.566 & 0.148 & 0.915 & 0.714 & 0.144 & 0.892 & 0.21 & 0.131 & 0.973\\
    0.17 & 0.692 & 0.137 & 0.873 & 0.673 & 0.143 & 0.885 & 0.815 & 0.139 & 0.864 & 0.26 & 0.157 & 0.968\\
    0.18 & 0.768 & 0.144 & 0.849 & 0.741 & 0.137 & 0.859 & 0.872 & 0.137 & 0.84 & 0.302 & 0.14 & 0.964\\

    \bottomrule

  \end{tabular}
  }
\end{table}

\begin{figure}[t]
    \centering
    \includegraphics[width=0.75\textwidth]{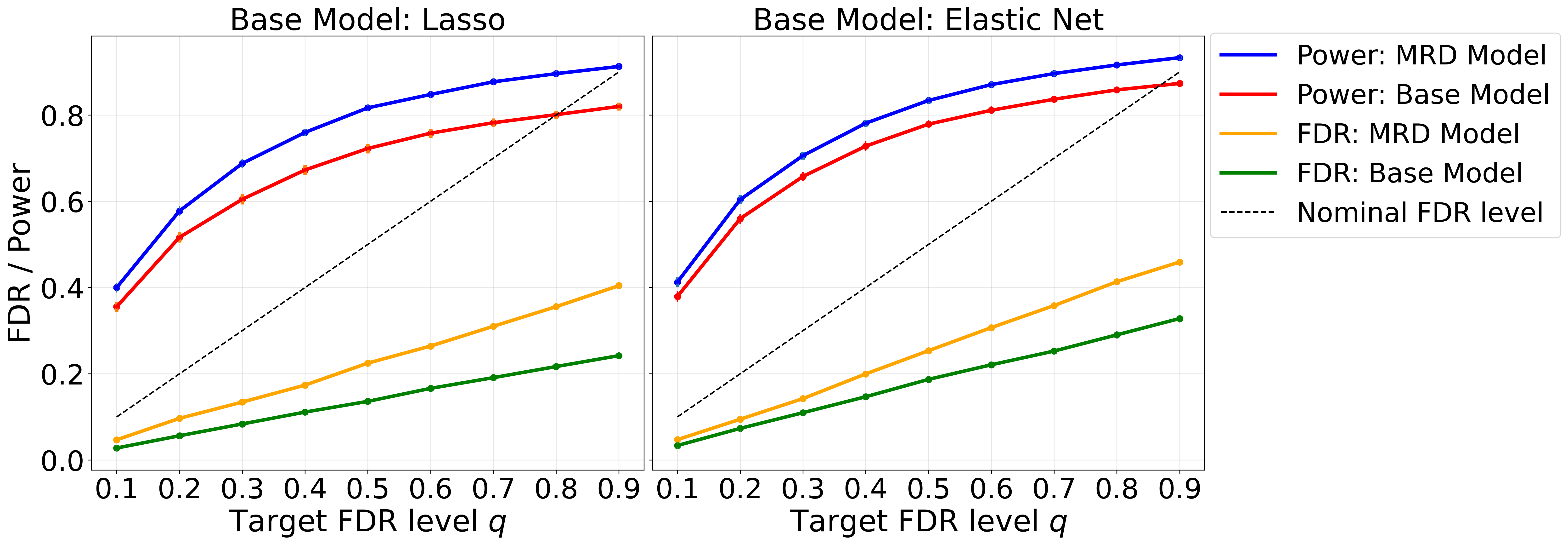}
    \caption{Synthetic experiments with correlated multivariate Gaussian features and a simulated response that follows a linear model with a constant signal strength $c=0.12$. Each graph displays the empirical FDR and power (evaluated by averaging over 100 independent experiments) as a function of the target FDR level $q$. Left: MRD lasso and lasso. Right: MRD elastic net and elastic net.} 
    \label{fig:lin_q}
\end{figure}

\begin{figure}[t]
    \centering
    \includegraphics[width=0.9\textwidth]{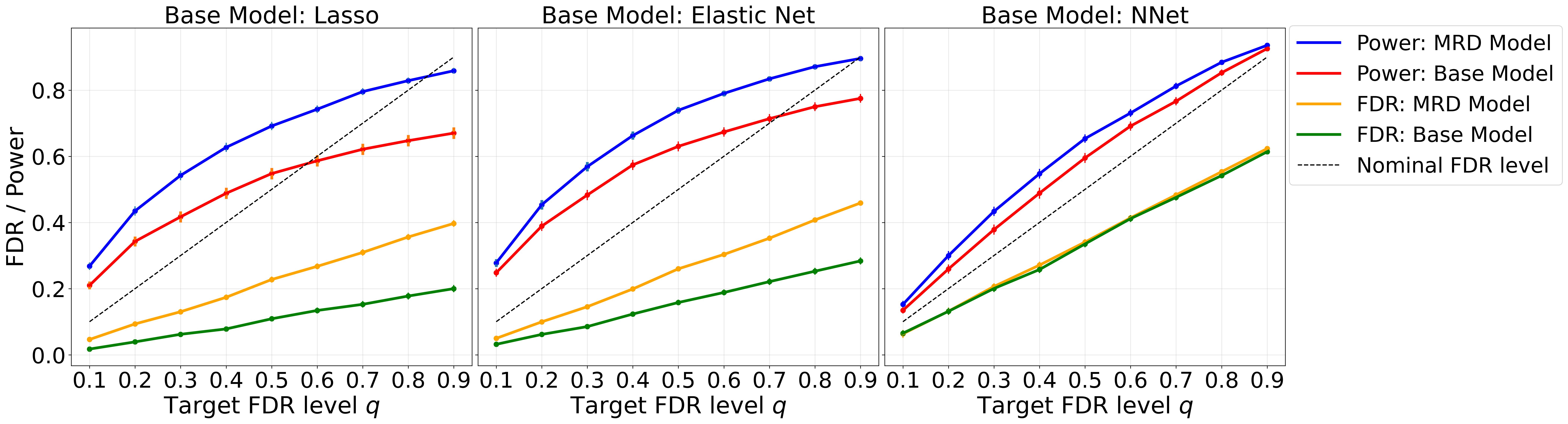}
    \caption{Synthetic experiments with correlated multivariate Gaussian features and a simulated response that follows a non-linear model. The signal strength is fixed and equals to $c=0.14$. Left: MRD lasso and lasso. Middle: MRD elastic net and elastic net. Right: MRD neural network (NNet) and NNet. The other details are as in Figure~\ref{fig:lin_q}.} 
    \label{fig:non_lin_q}
\end{figure}

\begin{figure}[t]
    \centering
    \includegraphics[width=0.65\textwidth]{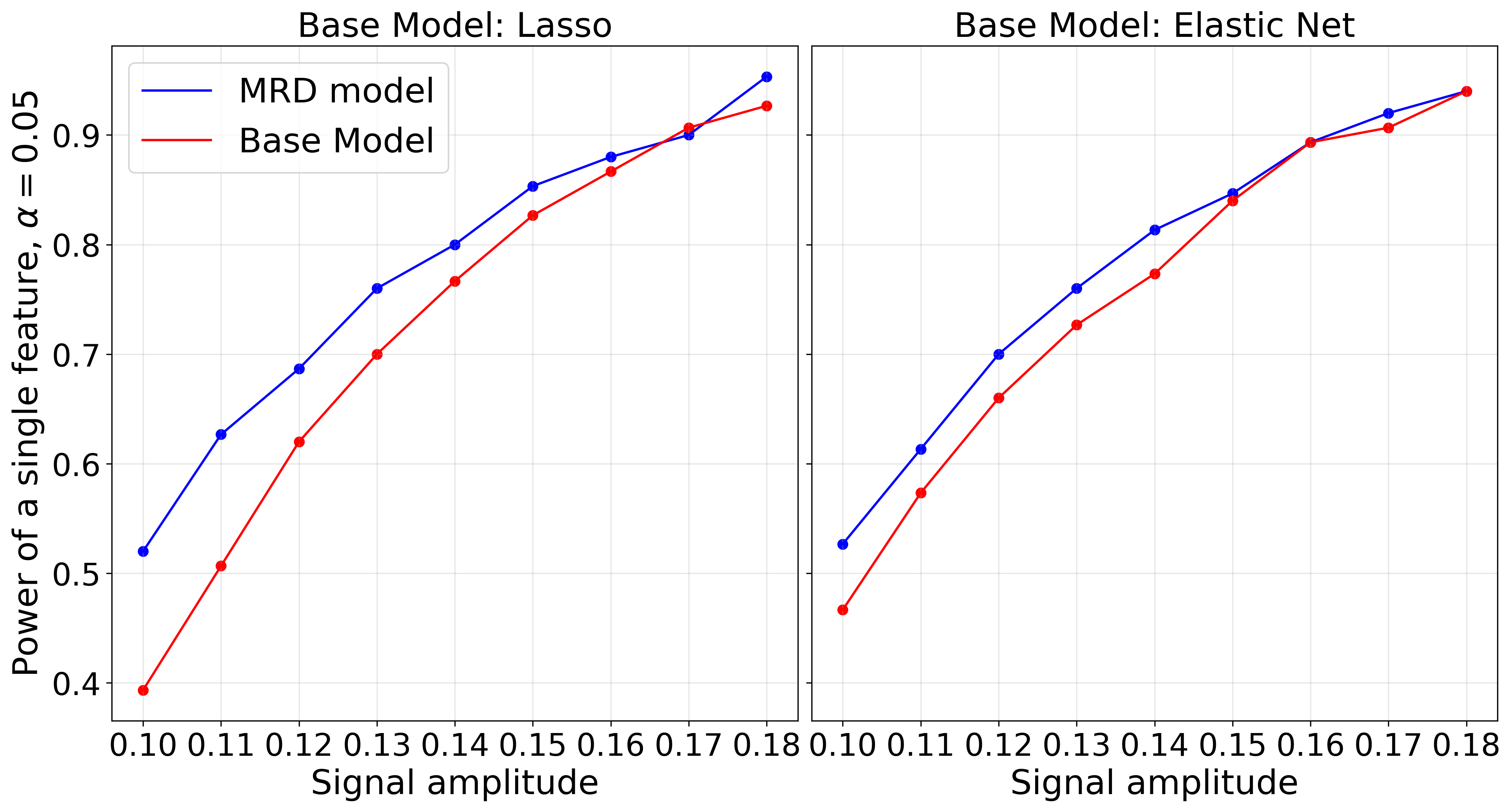}
    \caption{Testing for a single hypothesis on synthetic data with correlated multivariate Gaussian features and a simulated response that follows a linear model. The significance level is fixed and equals to $\alpha=0.05$. The empirical power is presented as a function of the signal strength $c$, and evaluated by averaging over 100 independent experiments. Left: MRD lasso and lasso. Right: MRD elastic net and elastic net.} 
    \label{fig:lin_one_hypo}
\end{figure}

\begin{figure}[t]
    \centering
    \includegraphics[width=0.9\textwidth]{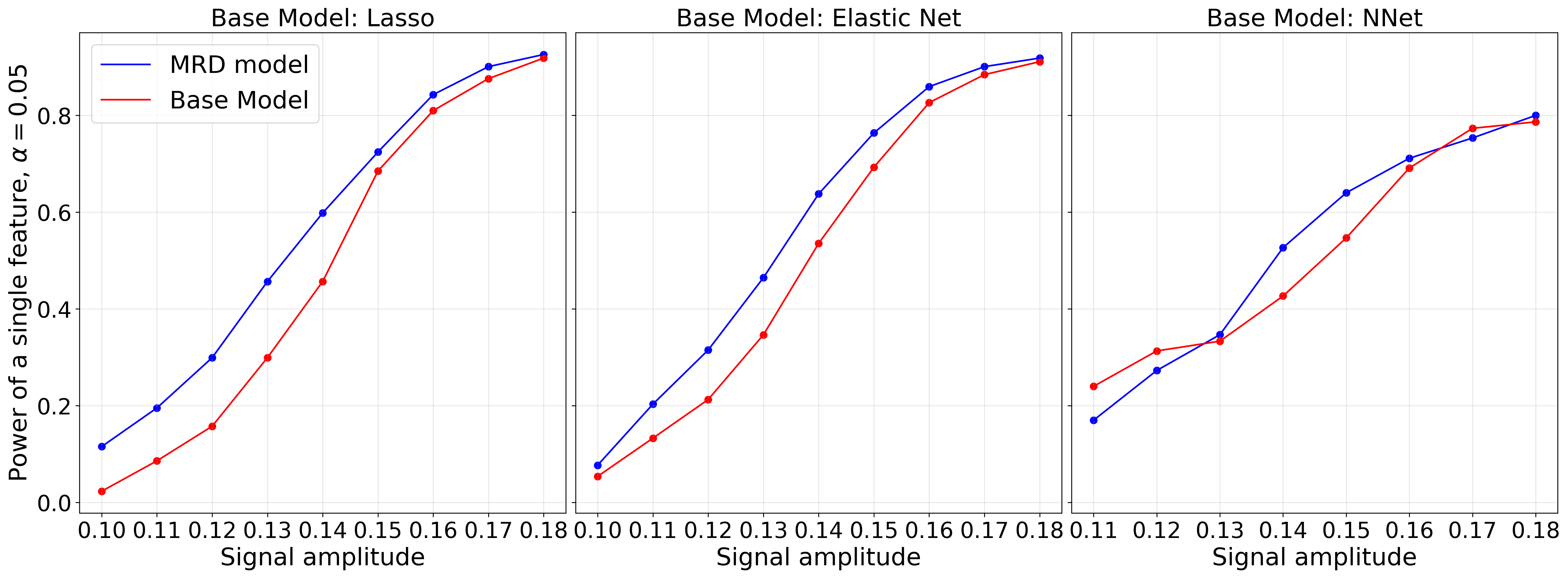}
    \caption{Testing for a single hypothesis on synthetic data with correlated multivariate Gaussian features and a simulated response that follows a non-linear model. Left: MRD lasso and lasso. Middle: MRD elastic net and elastic net. Right: MRD neural network (NNet) and NNet. The other details are as in Figure~\ref{fig:lin_one_hypo}.} 
    \label{fig:non_lin_one_hypo}
\end{figure}

\begin{figure}[t]
    \centering
    \includegraphics[width=0.65\textwidth]{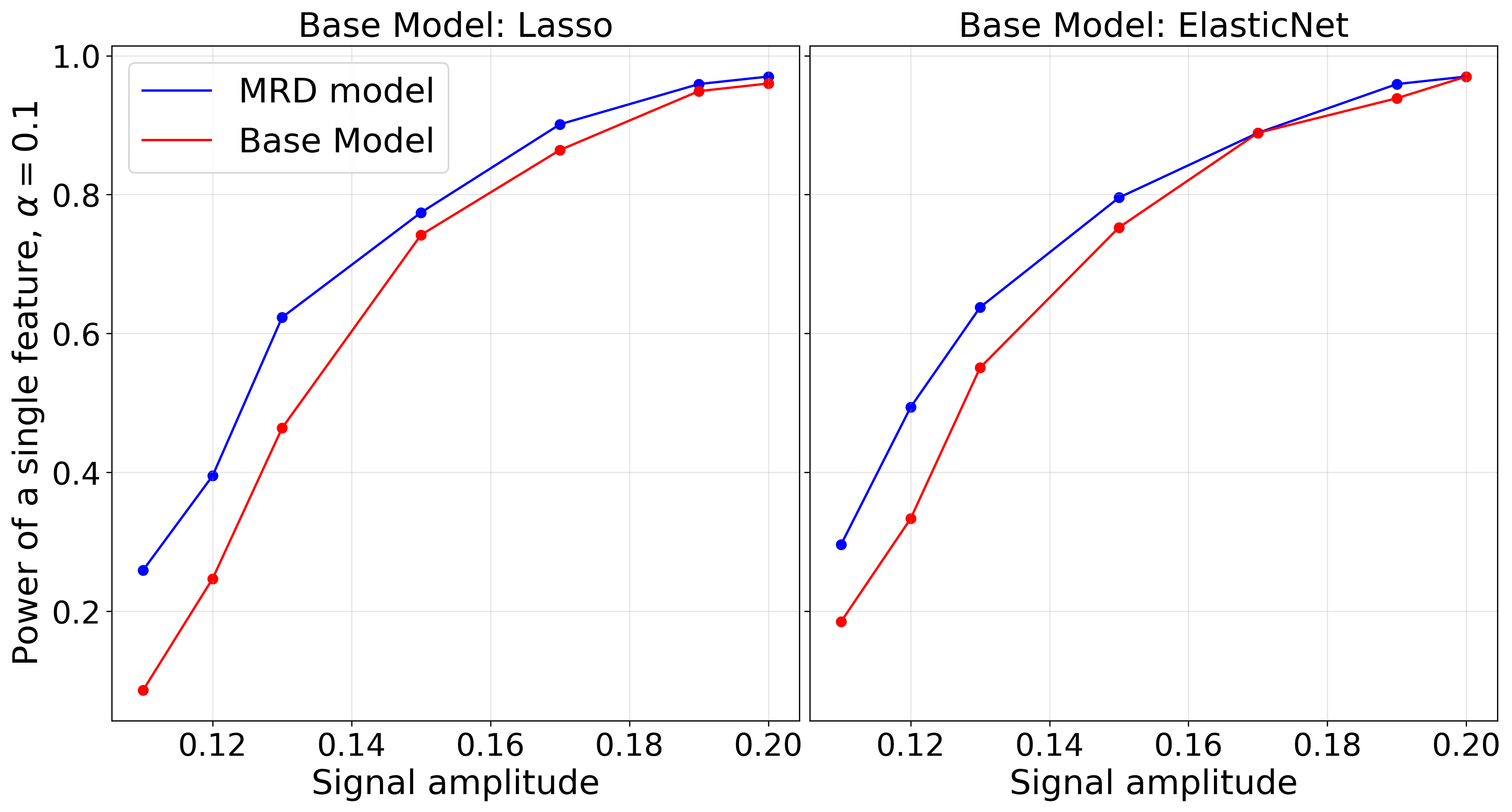}
    \caption{{Optimizing and testing for a single hypothesis on synthetic data with correlated multivariate Gaussian features and a simulated response that follows a non-linear model. The significance level is fixed and equals to $\alpha=0.1$. The empirical power is presented as a function of the signal strength $c$, and evaluated by averaging over 100 independent experiments. Left: MRD lasso and lasso. Right: MRD elastic net and elastic net.}} 
    \label{fig:non_lin_one_hypo_opt}
\end{figure}

\paragraph{Power as a function of the target FDR
}
The experiments presented above focus on feature selection with a fixed target FDR level. Figures~\ref{fig:lin_q}-\ref{fig:non_lin_q} present additional comparisons in which we vary the target FDR level while keeping the signal amplitude fixed. Following Figure~\ref{fig:lin_q}, which corresponds to the linear case (i.e., $Y \mid X$ follows \textbf{M1}), we can see that the MRD models outperform their baseline counterparts in terms of power. Observe that the FDR is controlled in all cases, where the base models are more conservative than their MRD versions. A similar trend is illustrated in Figure~\ref{fig:non_lin_q} that represents the nonlinear setting (where $Y \mid X$ follows \textbf{M2}). Here, we include a neural network model in addition to the sparse linear methods. Notice that the MRD neural network model attains better power than the baseline method, while the empirical FDR of the two is similar, both fall below the target level.

\begin{figure}[t]
    \centering
    \includegraphics[width=0.65\textwidth]{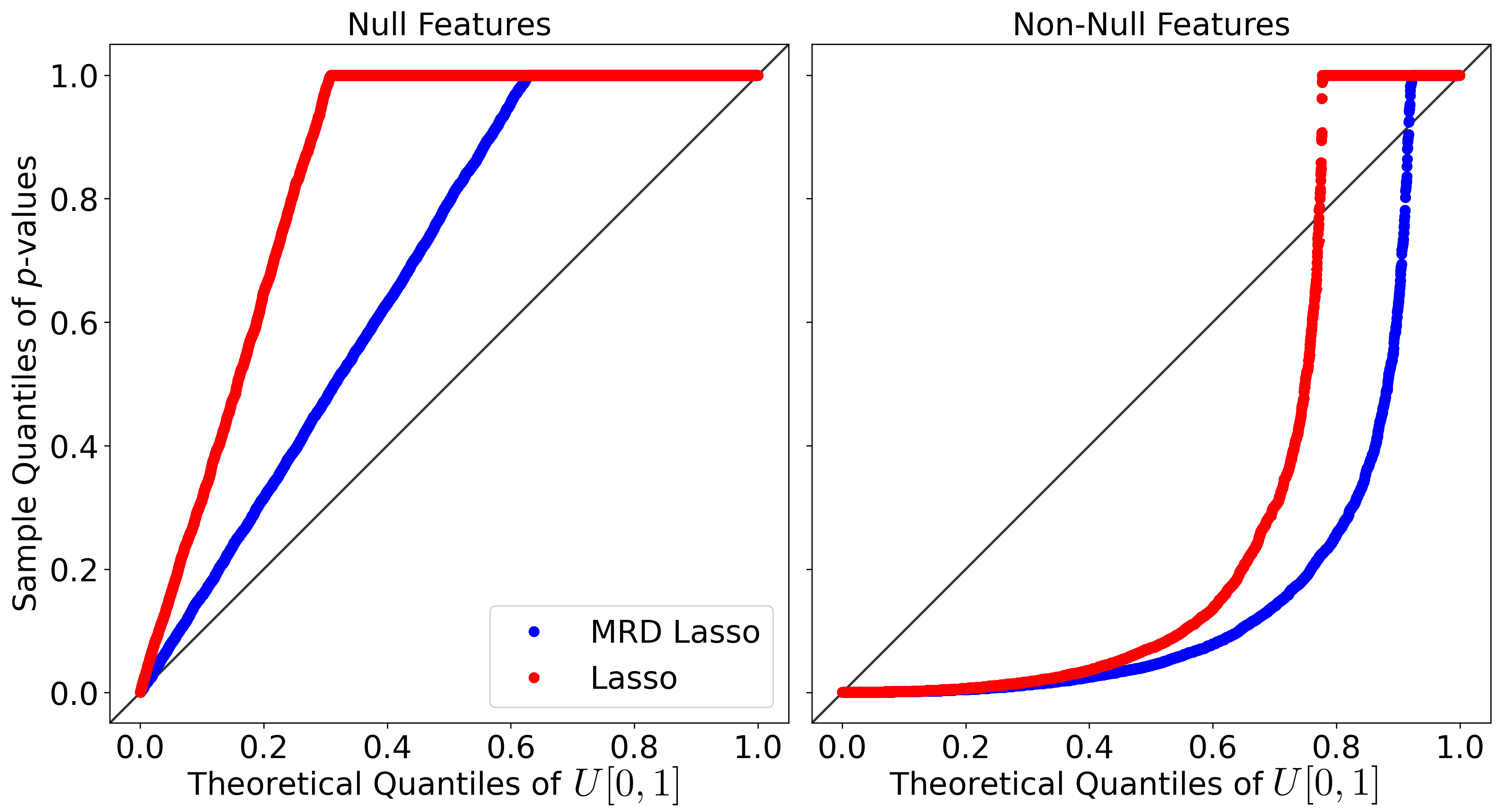}
    \caption{{Q-Q plots of the p-values produced in the experiment from Table~\ref{tab:non_linear_sin_sim_0.2} compared to the uniform distribution $U[0,1]$. Left: p-values correspond to the null features. Right: p-values correspond to the non-null features.}} 
    \label{fig:qqplots}
\end{figure}

\paragraph{p-value analysis}
{Herein, we study the behaviour of the p-values produced by the MRD approach. Figure~\ref{fig:qqplots} presents Q-Q plots of the p-values from the experiments summarized in Table~\ref{tab:non_linear_sim_0.2}, focusing on the case where $c=0.14$ with lasso as the baseline model. The left panel of Figure~\ref{fig:qqplots} presents the empirical quantiles of the p-values that correspond to the null features as a function of the quantiles of the uniform distribution on the $[0, 1]$ segment. As can bee seen, the HRT p-values for the null features are valid but conservative both for lasso and MRD lasso where the distribution of the latter p-values are closer to the uniform distribution. The right panel of Figure~\ref{fig:qqplots} presents an analogous plot but for the non-null features. Here, when combining HRT with MRD lasso we get p-values with lower quantiles that the ones of lasso, and both lie below the quantiles of the uniform distribution. Importantly, the MRD lasso leads to smaller p-values than lasso for the non-null features, demonstrating the advantage of the MRD approach.}

\subsection{Testing for a single hypothesis}
\label{sec:single_hyp}

We now analyze the effect of the MRD approach when testing for a single non-null hypothesis. Specifically, we choose at random a non-null $j$ from $\mathcal{H}_1$, and test for $H_{0,j}$ at level $\alpha=0.05$. 
We first consider the p-value evaluated via the HRT, using the same models and data from Tables~\ref{linear_sim_0.2_full}-\ref{non_linear_sim_0.2_full}. Figure~\ref{fig:lin_one_hypo} focuses on the case for which $Y \mid X$ follows a linear model (i.e., \textbf{M1}), presenting the power of lasso, elastic net, and their MRD versions as a function of the signal strength $c$. As can be seen, the results are aligned with the FDR experiments in the sense that the MRD approach increases the power of the test. Figure~\ref{fig:non_lin_one_hypo} repeats the same experiment in the non-linear case (where $Y \mid X$ follows \textbf{M2}). Here, once more, an improvement in power is achieved by the MRD approach. Notice that the gain in performance is not consistent for the base neural network model.

{Next, we demonstrate the advantage of the MRD when optimizing for a specific feature $j$. To this end,  we fit a model via Algorithm~\ref{alg:general_optimization} but set $\mathcal{P} = \{j\}$ in line 2 of Algorithm~\ref{alg:general_optimization}. We use the same synthetic experimental setting with a non linear data model (\textbf{M2}) as in the experiments from Section~\ref{sec:signal_strength}. Specifically, we fit MRD lasso and MRD elastic net for a single arbitrary non-null feature instead of all the features simultaneously. Figure~\ref{fig:non_lin_one_hypo_opt} presents the power of lasso, MRD lasso, elastic net, and MRD elastic net as a function of the signal strength $c$. As can be seen, the power of the MRD models is higher than that of the baseline models.}

\subsection{Additional experiment with interaction model}
\label{supp:interaction}
In this experiment, we set the auto-correlation parameter $\rho=0.25$ and choose the $Y|X$ to be $Y = \sum_{j=1}^{15} { X_{2j}X_{2j-1} } +\epsilon$ (i.e., $Y \mid X$ follows \textbf{M2}), such that in total there are 30 non-null elements out of total $d=100$ features. Figure~\ref{fig:inters_NN} presents the empirical power and FDR (target level of $q = 0.2$) obtained by a base model formulated as a neural network. We use the same network architecture and training procedure as described in Section~\ref{supp:models}, but with total $n=4000$ observations ($m=2000$), learning-rate of $0.006$, fit the model to a total of $200$ epochs, and set the CancelOut regularization parameter to $0.02$. (In this experiment, linear predictive models attain zero power, and hence omitted.)

\subsection{Experiments with sum of sines model}
\label{supp:link_func_}

In this section, we provide additional experiment with a different conditional modeling $Y \mid X$. Here, we use the same synthetic experiment setting as in the experiments with varying signal strength from Section~\ref{sec:signal_strength}, however generate $Y \mid X$ that follows \textbf{M3}. Table~\ref{tab:non_linear_sin_sim_0.2} presents the empirical power and FDR ($q$ = 0.2) obtained when using the elastic-net and lasso as the base predictive models. Following that table, the MRD achieves a gain in power compared to the base model, and the FDR is controlled.

\begin{table}[t]
    \caption{Synthetic experiments with correlated multivariate Gaussian features and a simulated response that follows a non-linear model with $g(X) = \sum_{j \in \mathcal{H}_1} {\sin{(X_j\beta_j)}}$. All standard errors are below 0.02. The other details are as in Table~\ref{tab:non_linear_sim_0.2}.}
     \label{tab:non_linear_sin_sim_0.2}
\resizebox{\textwidth}{!}{
  \centering
  \ra{1}

  \begin{tabular}{cllllcllllc}
    \toprule
    &\multicolumn{2}{c}{MRD Lasso}&
    \multicolumn{2}{c}{Lasso}& \multicolumn{1}{c}{\% imp. of} &
    \multicolumn{2}{l}{MRD Elastic Net}&
    \multicolumn{2}{c}{Elastic Net} & \multicolumn{1}{c}{\% imp. of}
    \\
    
    \cmidrule(r){2-3}
    \cmidrule(lr){4-5}
    \cmidrule(r){7-8}
    \cmidrule(r){9-10}

    $c$ &Power     & FDR     & Power  & FDR & power & Power  & FDR & Power  & FDR   & power\\
    \midrule
    0.08 & 0.143 & 0.066 & 0.081 & 0.023 &76.5& 0.153 & 0.075 & 0.113 & 0.055 & 35.4\\
    0.10 & 0.324 & 0.088 & 0.233 & 0.042 &39.1& 0.375 & 0.103 & 0.287 & 0.070 &30.7\\
    0.12 & 0.550 & 0.106 & 0.486 & 0.074 &13.2& 0.577 & 0.123 & 0.524 & 0.074 &10.1\\
    0.14 & 0.731 & 0.095 & 0.702 & 0.060 &4.1& 0.753 & 0.105 & 0.715 & 0.074 &5.3\\
    0.16 & 0.854 & 0.092 & 0.838 & 0.063 &1.9& 0.856 & 0.102 & 0.842 & 0.070 &1.7\\
    0.18 & 0.919 & 0.090 & 0.914 & 0.069 &0.5& 0.917 & 0.097 & 0.915 & 0.071 &0.2\\
    \bottomrule
  \end{tabular}
  }
\end{table}

\subsection{Experiments under model misspecification}
\label{sec:misspec_lin}

A limitation of the model-X randomization test is that $P_X$ must be known in order to sample the dummy features $\tilde{X}_j$, which are also used to formulate our RD loss. In this section, we demonstrate that the feature selection pipeline applied with the proposed MRD method is fairly robust to unknown or estimated distributions. For ease of computations, we conduct our experiments by choosing lasso or elastic net as the base models, fitted in the same way described in Section~\ref{supp:models}.

\begin{figure}
    \centering
    \includegraphics[width=0.78\textwidth]{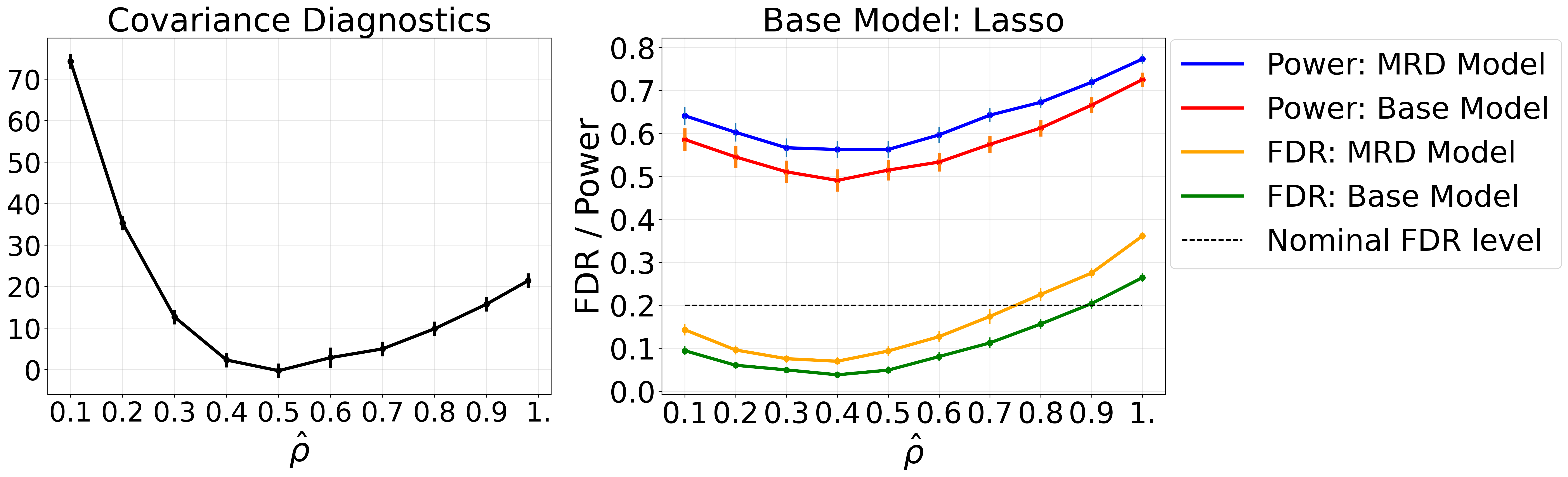}
    \caption{Experiment with dummy features whose quality varies with $\hat{\rho}$. Here, the true $P_X$ follows a multivariate Gaussian distribution with with auto-correlation $\rho=0.5$. Left: sum of covariance diagnostics over all $j=1,\dots,p$ (lower values correspond to more accurate dummies) as a function of $\hat{\rho}$. Right: empirical power and FDR.} 
    \label{fig:rob_gof}
\end{figure}

\paragraph{Varying the estimation quality of the conditionals}
We begin with an experiment that demonstrates the effect of the density estimation on the empirical FDR. We generate features as in Section~\ref{sec:syn_exp}, but with an auto-correlation coefficient $\rho=0.5$, higher than the one used in the synthetic experiments from that section. Here, the model of $Y \mid X$ follows \textbf{M2} with a fixed signal strength $c=0.16$, where we use lasso as a base predictive model. To control the quality of the estimation of $P_X$, we sampled the dummy features $\tilde{X}_j$ from $Q_{X_j \mid X_{-j}}$, where $Q_X$ also follows a multivariate Gaussian however with an auto-correlation coefficient $\hat{\rho}$ that can differ from the true $\rho=0.5$. We evaluate the quality of the dummies $\tilde X_j \sim Q_{X_j \mid X_{-j}}$ using the covariance goodness-of-fit diagnostic, presented by \citealp[Section 5]{DeepKnockoffs}. In short, this diagnostic presents an unbiased estimate of the distance between $\mathrm{Cov}(X_j,X_{-j})$ and $\mathrm{Cov}(\tilde{X}_j,X_{-j})$, where larger values imply worse estimation of the conditional. The left panel of Figure~\ref{fig:rob_gof} depicts the sum of the covariance diagnostics over $j=1,\dots,p$. Each point in that graph is the average value taken over 50 independent data sets, where each is of size $10,000$ samples. As can be seen, the value of the sum covariance diagnostics is close to zero when $\hat{\rho} = 0.5$, as expected, and gets higher as $\hat{\rho}$ is far from the true value of $\rho=0.5$.

Next, we turn to study the effect of sampling inaccurate dummies on the performance of our proposed method. To this end, we use the inaccurate dummies (sampled from $Q_{X_j \mid X_{-j}}$) both for fitting the MRD model and for computing the test p-values. The right panel of Figure~\ref{fig:rob_gof} depicts the empirical power and FDR of both MRD lasso and its base version, evaluated on 50 independent data sets of size $n=1000$ each.
 Following that figure, we observe that the MRD lasso is slightly more sensitive to the use of inaccurate dummy features, however a violation in the FDR control occurs only when $\hat{\rho} \geq 0.8$ for which the sum of the covariance diagnostics is above 10. This experiment illustrates that our method is fairly robust to inaccurate estimation of $P_X$, where invalid results may be obtained when this estimation is of poor quality. Importantly, this statement holds true for all model-X randomization tests, as the generated p-values are valid only when $P_X$ is known. In terms of power, the MRD lasso presents, once again, a solid gain in performance compared to lasso when $\hat{\rho}=\rho$, where the FDR is controlled. (Recall that the correlation structure here is more significant than the synthetic experiments from Section~\ref{sec:syn_exp}, in which $\rho \in \{0.1,0.25\}$.)

\paragraph{Multivariate Gaussian} Consider the experimental setup from the experiments with varying signal strength in Section~\ref{sec:syn_exp}, where $X \sim \mathcal{N}(\mu,\Sigma)$ is drawn from an auto-regression model. There, we fit the MRD models and apply the HRT by drawing $\tilde{X}_j$ from the true $P_{X_j \mid X_{-j}}$. Here, we conduct similar experiments, but sample $\tilde{X}_j$ from $\hat{P}_{X_j \mid X_{-j}}$, where $\hat{P}_X \sim \mathcal{N}(\hat{\mu},\hat{\Sigma})$ with mean $\hat{\mu}$ and covariance matrix $\hat{\Sigma}$ that are estimated from the observed data. Following Figures~\ref{fig:linear_est_params}-\ref{fig:non_linear_est_params}, we can see that all predictive models---including ours---result in FDR below the nominal 20\% level, both in the linear and non-linear settings. Notice that our MRD approach consistently improves the power of each base model, which is in line with the experiments from Section~\ref{sec:syn_exp}.

\begin{figure}
    \centering
    \includegraphics[width=0.75\textwidth]{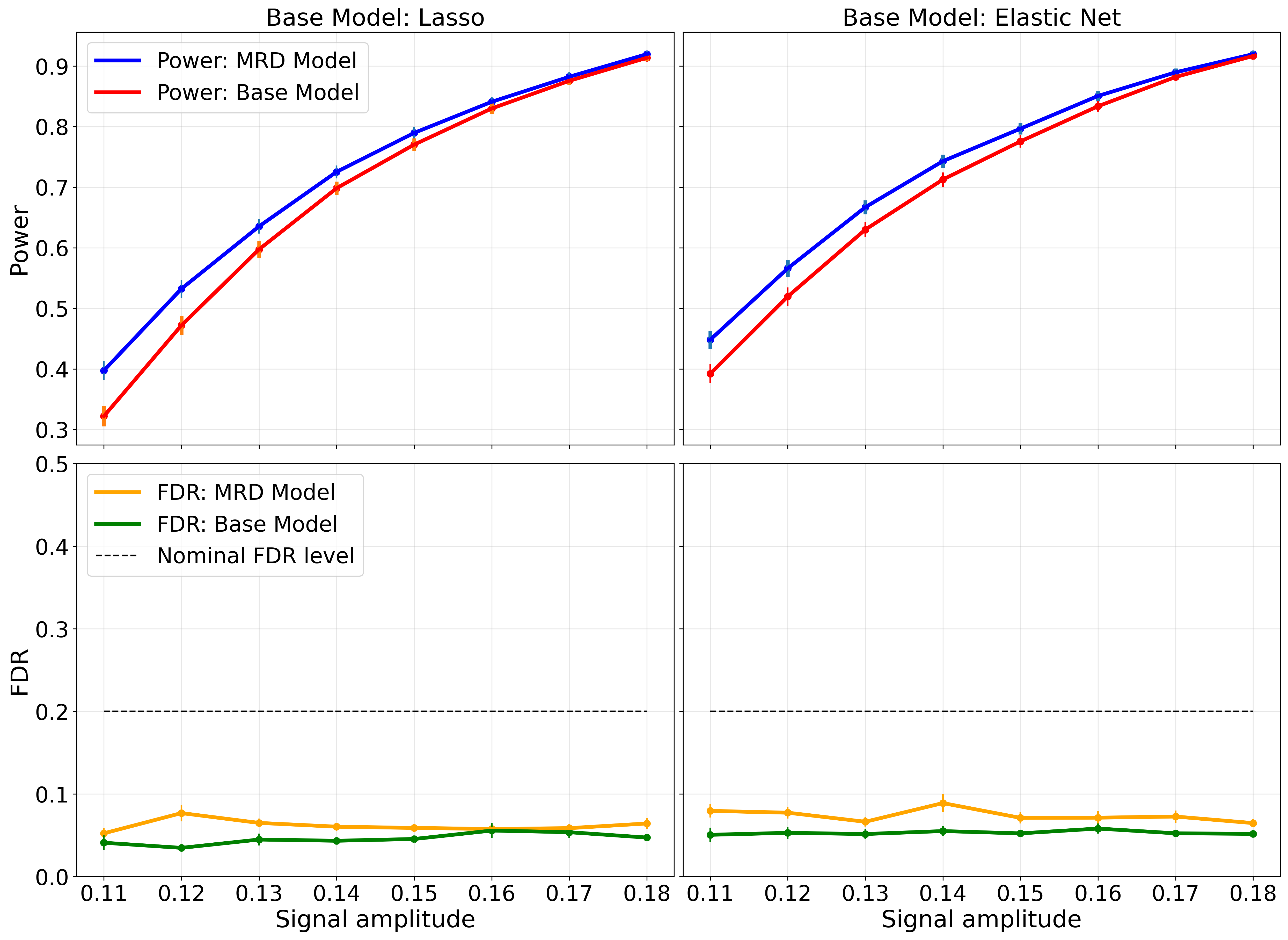}
    \caption{Robustness experiments with correlated multivariate Gaussian features and a simulated response that follows a linear model with varying signal strength $c$. The mean and covariance of $P_X$ are  estimated from the data. Left: power (top) and FDR (bottom) of MRD lasso and lasso. Right: power (top) and FDR (bottom) of MRD elastic net and elastic net. The performance metrics are averaged over 100 independent experiments.} 
    \label{fig:linear_est_params}
\end{figure}

\begin{figure}
    \centering
    \includegraphics[width=0.75\textwidth]{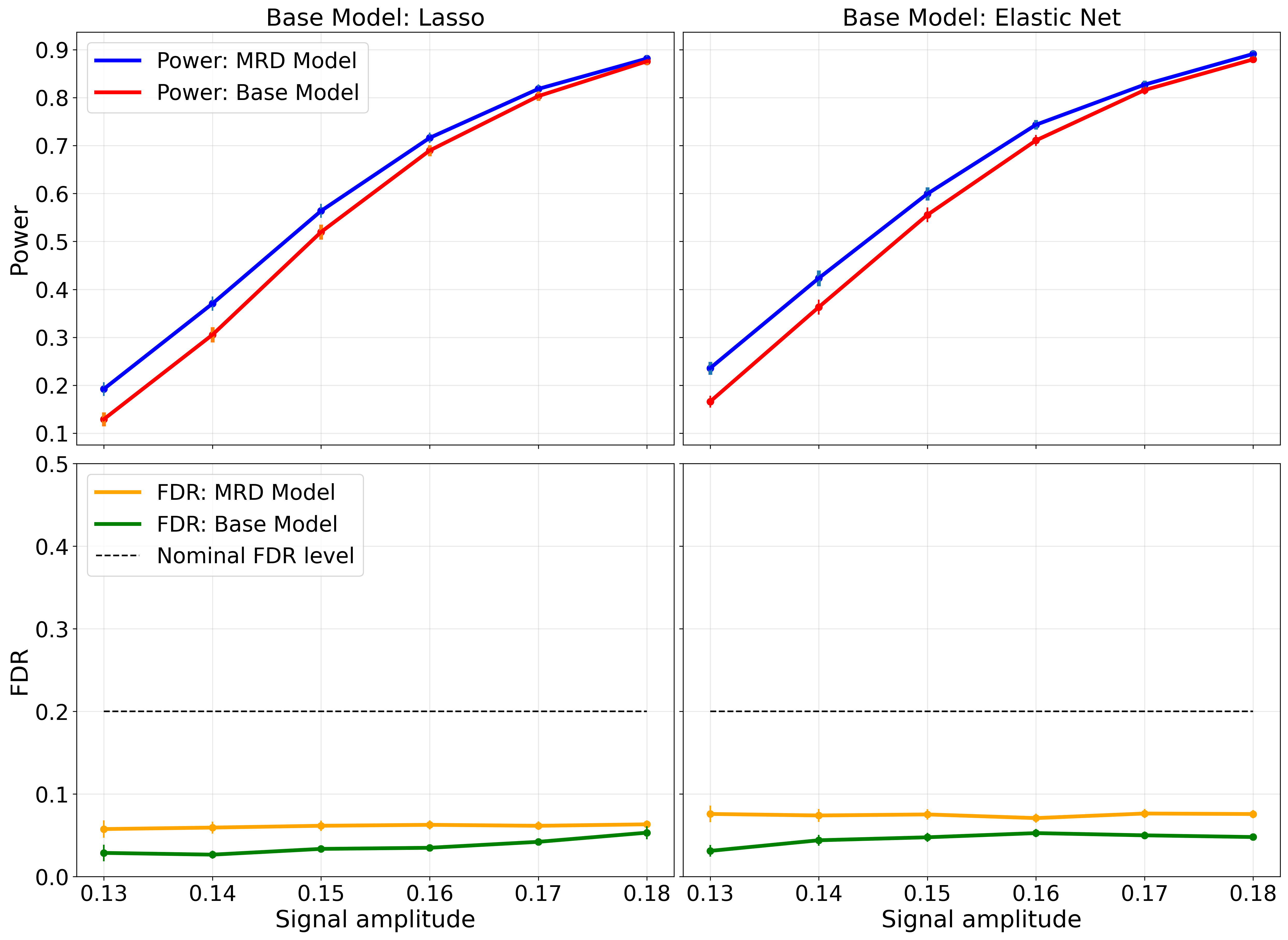}
    \caption{Robustness experiments with correlated multivariate Gaussian features and a simulated response that follows a non-linear model with varying signal strength $c$. The other details are as in Figure~\ref{fig:linear_est_params}.} 
    \label{fig:non_linear_est_params}
\end{figure}

\begin{figure}
    \centering
    \includegraphics[width=0.75\textwidth]{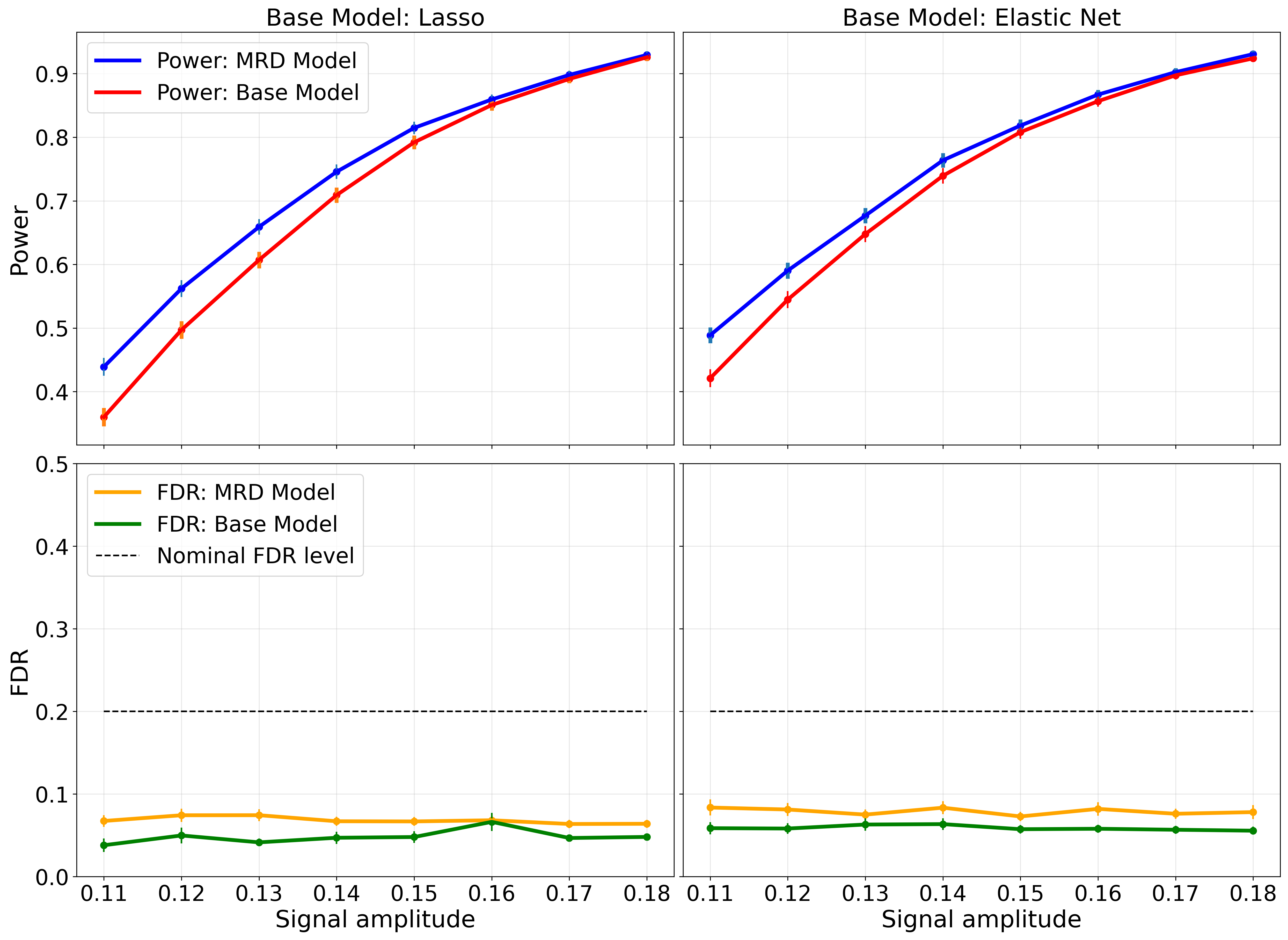}
    \caption{Robustness experiments with correlated multivariate Gaussian mixture features and a simulated response that follows a linear model with varying signal strength $c$. The other details are as in Figure~\ref{fig:linear_est_params}.} 
    \label{fig:linear_gmm}
\end{figure}

\begin{figure}
    \centering
    \includegraphics[width=0.75
    \textwidth]{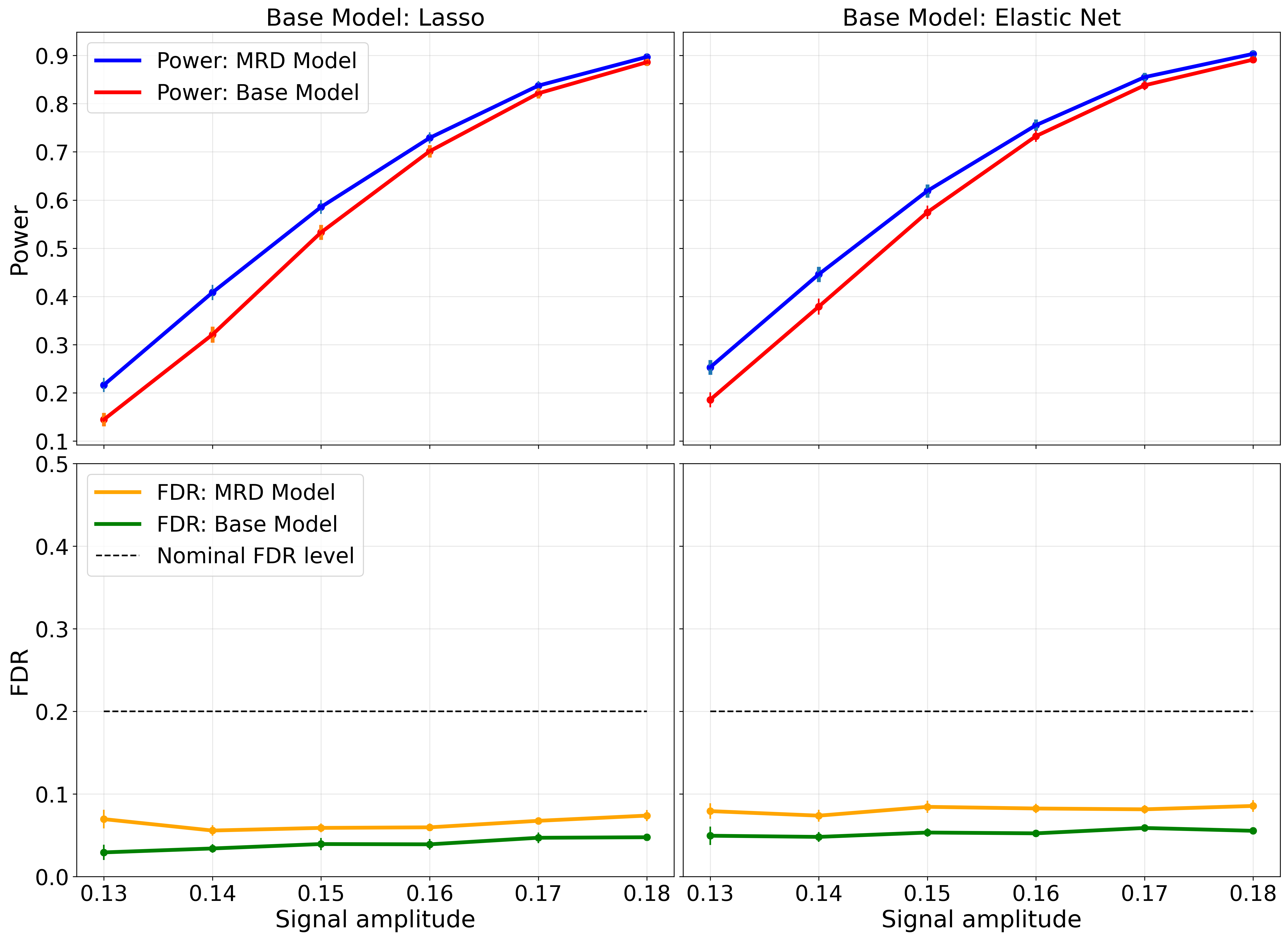}
    \caption{Robustness experiments with correlated multivariate Gaussian mixture features and a simulated response that follows a non-linear model with varying signal strength $c$. The other details are as in Figure~\ref{fig:linear_est_params}.} 
    \label{fig:non_linear_gmm}
\end{figure}

\paragraph{Gaussian mixture model} We now present a more challenging experiment in which $X$ is drawn independently from a multivariate Gaussian mixture model:
\begin{equation}
    X \sim \Bigg\{\begin{matrix} \mathcal{N}(0,\Sigma_1), \quad \mathrm{w.p} \quad \frac{1}{3}, \\
    \mathcal{N}(0,\Sigma_2), \quad \mathrm{w.p} \quad \frac{1}{3}, \\
    \mathcal{N}(0,\Sigma_3), \quad \mathrm{w.p} \quad \frac{1}{3}.  \end{matrix}
\end{equation}
Above, $\Sigma_1,\Sigma_2$ and $\Sigma_3$ have the same correlation structure described in Section~\ref{sec:syn_exp}, with auto-correlation coefficients $\rho_1 = 0.1$, $\rho_2 = 0.2$, and $\rho_3 = 0.3$, respectively. Similar to the experiments presented above, we assume that $X$ follows a multivariate Gaussian and we estimate its parameters from the observed data. This allows us to study the robustness of the proposed MRD approach to model misspecification. We follow the experimental protocol described in the experiments with varying signal strength from Section~\ref{sec:signal_strength} (with \textbf{M1} for the non-linear setting), and summarize the results in Figures~\ref{fig:linear_gmm}-\ref{fig:non_linear_gmm}. As can be seen, the FDR obtained by all models is below the desired 20\% level both in the linear and non-linear settings, indicating the robustness of the feature selection procedure. Moreover, our MRD approach yields higher power than the base models, demonstrating that the proposed method is also effective when the distribution of $X$ is unknown.

\begin{figure}
    \centering
    \includegraphics[width=0.95
    \textwidth]{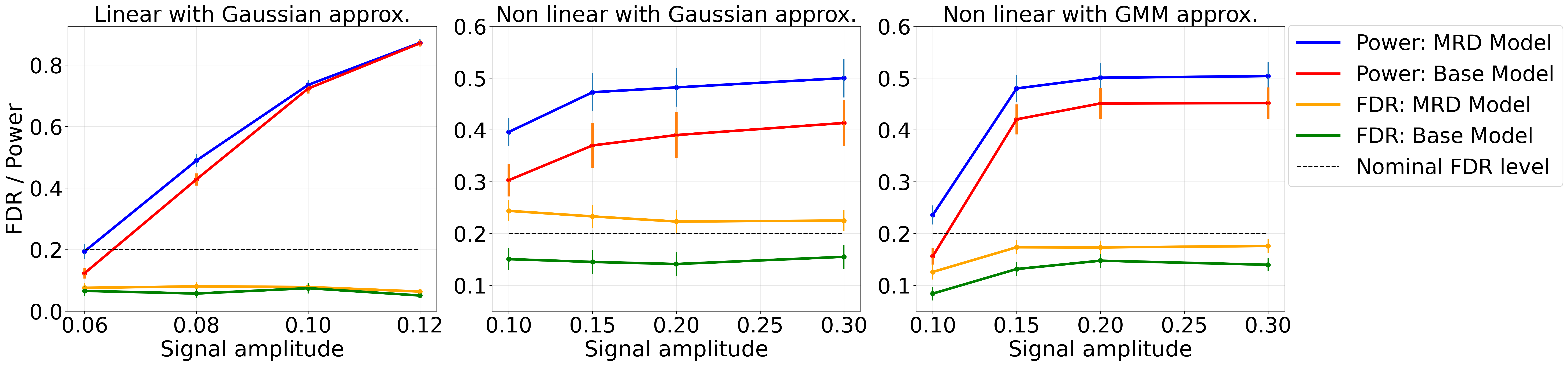}
    \caption{Experiments with correlated multivariate Student $t$-distribution features and a simulated response with varying signal strength $c$. Power and FDR are evaluated on 50 independent datasets. Left: response that follows a linear model, where the dummies are sampled from a naive multivariate Gaussian approximation of $P_X$. Middle: response that follows a non linear model, where the dummies are also sampled from a naive multivariate Gaussian. Right: response that follows a non linear model, where the dummies are sampled from a GMM approximation of $P_X$.} 
    \label{fig:student_t}
\end{figure}

\paragraph{Multivariate Student's $t$-distribution}
In the experiments from previous paragraphs, we use a simple Gaussian approximation of $P_X$, and we always obtain a control of the FDR. Now, we turn to a scenario where such an approximation does not result in FDR control, and demonstrate that using a better density estimation for $P_X$ addresses tackles this limitation. To this end, we generate $n=1000$ samples of $X$, where $X$ follows a multivariate Student's $t$-distribution with zero mean, covariance matrix with entries $\Sigma_{i,j} = \rho^{|i-j|}$ where $\rho=0.1$, and $\nu = 5$ degrees of freedom; refer to \cite[Section~6.5]{DeepKnockoffs} for the precise definition of this data generating function. We begin this experiment by sampling $Y$ that follows a linear model (i.e., \textbf{M1}) of a varying signal strength $c$, as described in Section~\ref{sec:syn_exp} of the main manuscript, where we use the same multivariate Gaussian approximation of $P_X$ as in the previous paragraphs. The left panel of Figure~\ref{fig:student_t} shows the empirical power and FDR evaluated on 50 independent data sets, comparing the performance of an elastic net model and its MRD version. Next, we generate the response $Y$ using the non linear model \textbf{M2}, and again vary the signal strength $c$ as described in Section~\ref{sec:signal_strength} of the main manuscript. As before, we approximate the distribution of $X$ by fitting a multivariate Gaussian on the observed data. The middle panel of Figure~\ref{fig:student_t} presents the empirical power and FDR of this non linear setting, demonstrating how the MRD elastic net fails to control the FDR while the base elastic net model is below the nominal FDR rate. This violation is due to the inccurate estimation of $P_X$ provided by the naive multivariate Gaussian fit. To alleviate this, we now turn to approximate $P_X$ using a Gaussian Mixture Model (GMM) \cite{GMM} with 3 components, where we use Python's \texttt{sklearn} package to fit such a GMM model. The right panel of Figure~\ref{fig:student_t} presents the empirical power and FDR obtained by sampling the dummy features from the fitted GMM model \cite{gimenez2019knockoffs}, instead of a naive multivariate Gaussian, where we use the same non-linear setting as in the middle panel. Following the right panel of that figure, we can see that the FDR is controlled across all signal strength values, showing that by using better techniques to approximate $P_X$ we get more reliable results. This is in line with many other techniques proposed in the model-X literature that offer practical solutions to sampling more accurate dummies, as discussed in Section~\ref{sec:unknown_p} of the main manuscript. Lastly, in terms of power, the one can see that the MRD models outperform their base versions across the board, emphasizing the superiority of our proposal.

\subsection{Experiments with varying auto-correlation parameter $\rho$}
\label{sec:varying_rho}
\begin{figure}
    \centering
    \includegraphics[width=0.95
    \textwidth]{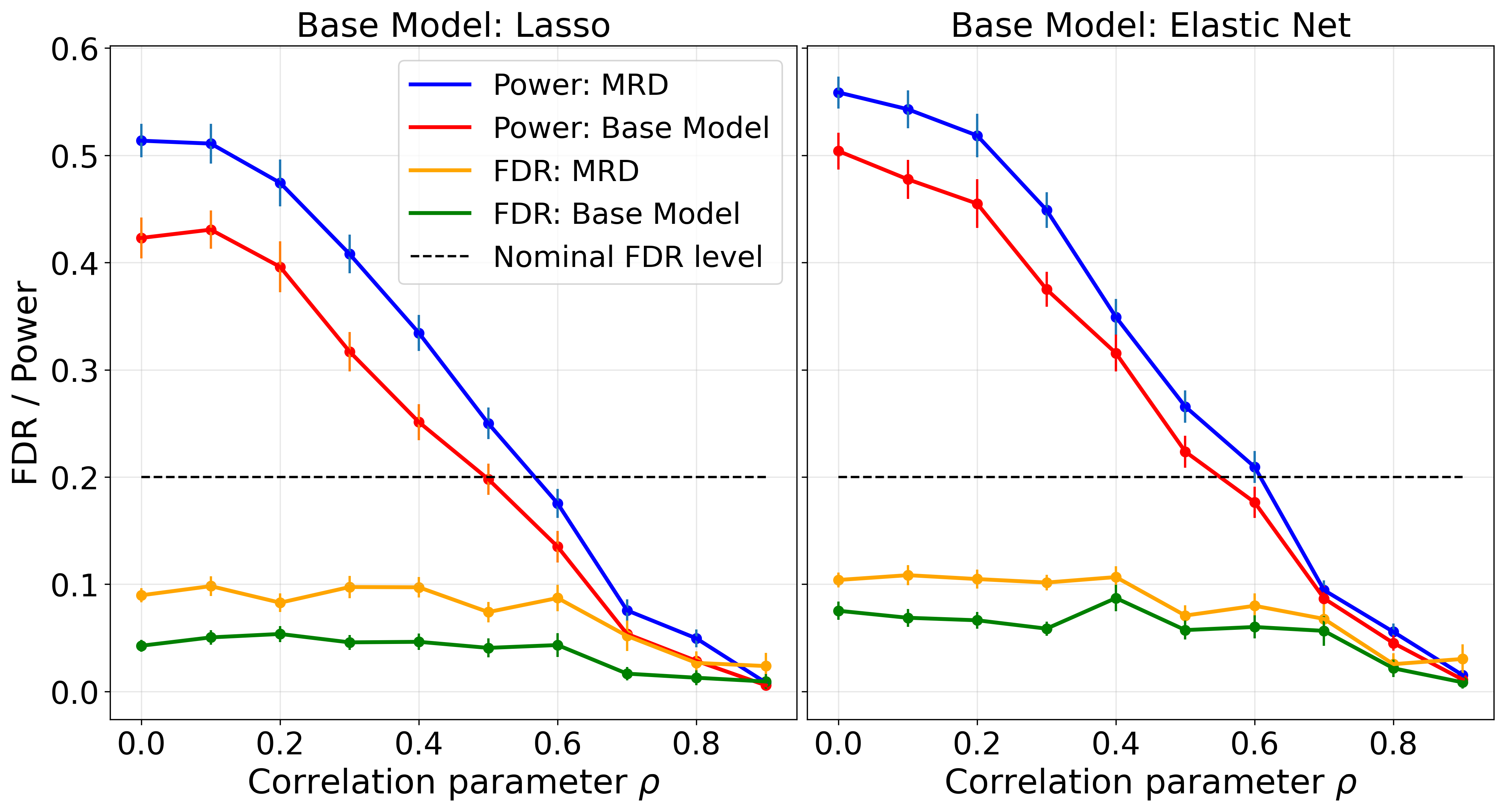}
    \caption{{Experiments with correlated features that follow $\mathcal{N}(0,\Sigma)$ where $\Sigma_{i,j}=\rho^{|i-j|}$ with varying auto-correlation parameter $\rho$, and a simulated response. Power and FDR are evaluated on 100 independent data sets.}} 
    \label{fig:varying_rho}
\end{figure}

In this section, we study the effect of the correlation between features on the proposed MRD approach. To this end, we follow the experimental setting described in Section~\ref{sec:syn_exp}, where $Y\mid X$ follows \textbf{M2} with a fixed signal strength $c=0.14$ and $m=400$, and vary the auto-correlation parameter $\rho$. Figure~\ref{fig:varying_rho} presents the empirical power and FDR, evaluated over 100 independent data sets, as a function of $\rho$ for lasso, elastic net and their MRD versions. As can be seen, the power of all methods decreases as $\rho$ increases, indicating the challenge of testing for conditional independence with highly correlated features. However, the MRD approach results in higher power for almost every $\rho$ while maintaining the FDR under control, demonstrating the advantage of the MRD approach even when working with correlated features.

\subsection{A comparison of CV-HRT to dCRT}
\label{sec:dcrt_compare}

{In this section, we use simulated data to compare the performance of CV-HRT~\cite[Algorithm~4]{HRT} to that of dCRT~\cite{dCRT} and of the model-X knockoffs \cite{Knockoffs}. (Recall that the former relies on cross-validating and thus avoids the naive data-splitting of the vanilla HRT procedure.)
To this end, we generate $X \in \mathbb{R}^{d}$ that follows an autoregression model with correlation coefficient $\rho=0.6$, and set $d=100$. Then, we generate a response variable $Y=(X^T\beta + \epsilon)^3$, where $\epsilon \sim \mathcal{N}(0,1)$ is a noise component, and $\beta \in \mathbb{R}^d$ is a sparse vector whose 30 first entries are equal to $c=1.5$; we set the rest 70 elements in $\beta$ to zero.  We choose the elastic net to be the base predictive model for both CV-HRT and its MRD version, and compare the performance of these two tests to dCRT and to the model-X knockoffs. The CV-HRT is implemented by setting the number of folds to be equal to $K=20$. We implemented the dCRT in Python, by following the official R package provided by the authors;\footnote{\url{https://github.com/moleibobliu/Distillation-CRT}} we run the dCRT method using the default choice of parameters, provided in the official R package. We run the model-X knockoffs with a second order estimation for generating the knockoffs, by using the package provided in \cite{DeepKnockoffs} with its default choice of parameters.\footnote{We use the software package from \url{https://github.com/msesia/deepknockoffs}}}

{Table~\ref{tab:dcrt_compare} compares the performance of the above four frameworks (averaged over 50 trials) for a varying number of samples $n$, where the target FDR level is fixed and equals to $q=0.1$.
Following that table, we can see that the gap between CV-HRT and the dCRT is relatively small, and the MRD approach closes this gap even further, achieving essentially the same power as dCRT. The model-X knockoffs achieves lower power than all other frameworks. In all experiments, the empirical FDR is controlled, where the MRD version of CV-HRT tends to have a lower empirical FDR compared to dCRT and the model-X knockoffs. Lastly, we note that CV-HRT requires fitting $K$ different models (20 in this experiment), less than the $d$ leave-one-covariate-out models that are required to implement the dCRT test. Therefore, in general, the computational complexity of CV-HRT may be lower than that of the dCRT.}

\begin{table}[t]
    \caption{{Synthetic experiments with correlated multivariate Gaussian features of dimension $d=100$, and a simulated response that follows a non-linear model with varying number of samples $n$. The right column represents the result of the dCRT procedure, and the rest represent the CV-HRT procedure with 20 folds. The empirical FDR (nominal level $q=0.1$) and power are evaluated by averaging over 50 independent experiments. All standard errors are below 0.02.}}
     \label{tab:dcrt_compare}
\fontsize{8}{8}\selectfont
  \centering
  \ra{1}

  \begin{tabular}{cllllllll}
    \toprule
    &\multicolumn{2}{c}{Lasso}&
    \multicolumn{2}{c}{MRD Lasso}&
    \multicolumn{2}{c}{dCRT}&
    \multicolumn{2}{c}{Knockoffs}
    \\
    
    \cmidrule(r){2-3}
    \cmidrule(r){4-5}
    \cmidrule(r){6-7}
    \cmidrule(r){8-9}

    $n$ &Power     & FDR     & Power  & FDR & Power  & FDR & Power  & FDR  \\
    \midrule
    600 & 0.24 & 0.02 & 0.25 & 0.03 & 0.25 & 0.06 & 0.13 & 0.05\\
    1000 & 0.50 & 0.02 & 0.52 & 0.01 & 0.52 & 0.05 & 0.42 & 0.05\\
    1400 & 0.70 & 0.02 & 0.72 & 0.02 & 0.72 & 0.05 & 0.59 & 0.08\\
    \bottomrule
  \end{tabular}
\end{table}

\section{Supplementary details on real data experiments}
\label{supp:real-data}
\subsection{Description of the data}

In Section \ref{sec:real-data} of the main manuscript, we analyze which genetic variations are linked to changes in human immunodeficiency virus drug-resistance. We download the data set from \url{http://hivdb.stanford.edu/pages/published_analysis/genophenoPNAS2006} and deploy the same pre-processing suggested in \cite{DeepKnockoffs}. Specifically, we selected $d=150$ features in total, where $75$ of them are chosen because they are already reported as important,\footnote{We use the database from \url{https://hivdb.stanford.edu/dr-summary/comments/PI}} and the rest $75$ are those who have the most frequently occurring mutations. After discarding samples that have missing values, we have 1555 examples with 150 features each.

\subsection{Semi-synthetic experiments with a non-linear model}
\label{supp:semi_non_linear}
\begin{figure}[t]
    \centering
    \includegraphics[width=0.9\textwidth]{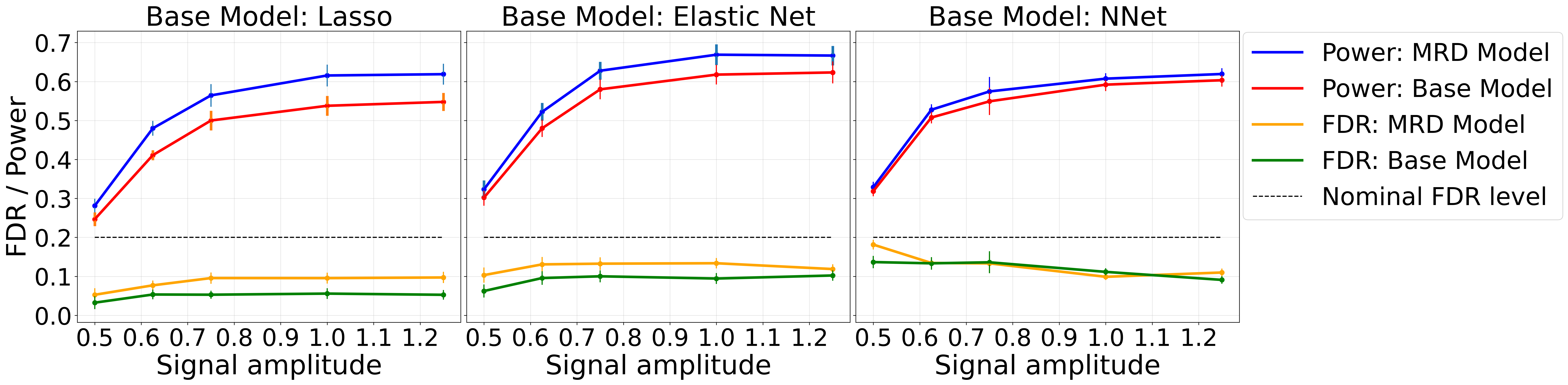}
    \caption{Semi-synthetic experiments with real HIV mutation features and a simulated response that follows a non-linear model with varying signal strength $c$. Empirical power and FDR with $q= 0.2$ are evaluated over $20$ random train/test splits of the data. The other details are as in Figure~\ref{fig:non_lin_semi}.} 
    \label{fig:non_lin_semi_with_NN}
\end{figure}

Here we provide the complementary results of the semi-synthetic non-linear data experiments, discussed in Section~\ref{sec:real-data}. Figure~\ref{fig:non_lin_semi_with_NN} presents the empirical power and FDR evaluated over 20 realizations of the dataset (random train/test splits of the data), using lasso, elastic net and neural network as the base models. In addition to the discussion in Section~\ref{sec:real-data}, it can be seen that the improvement of the neural network is consistent but less significant than those of the linear predictive models (lasso and elastic net). However, although we misspecify the distribution $P_X$, the empirical FDR is below the nominal level $q=0.2$ in the neural network as well as lasso and elastic net. 

\subsection{Semi-synthetic experiments with a linear model}
\label{supp:semi_linear}
\begin{figure}[t]
    \centering
    \includegraphics[width=0.7\textwidth]{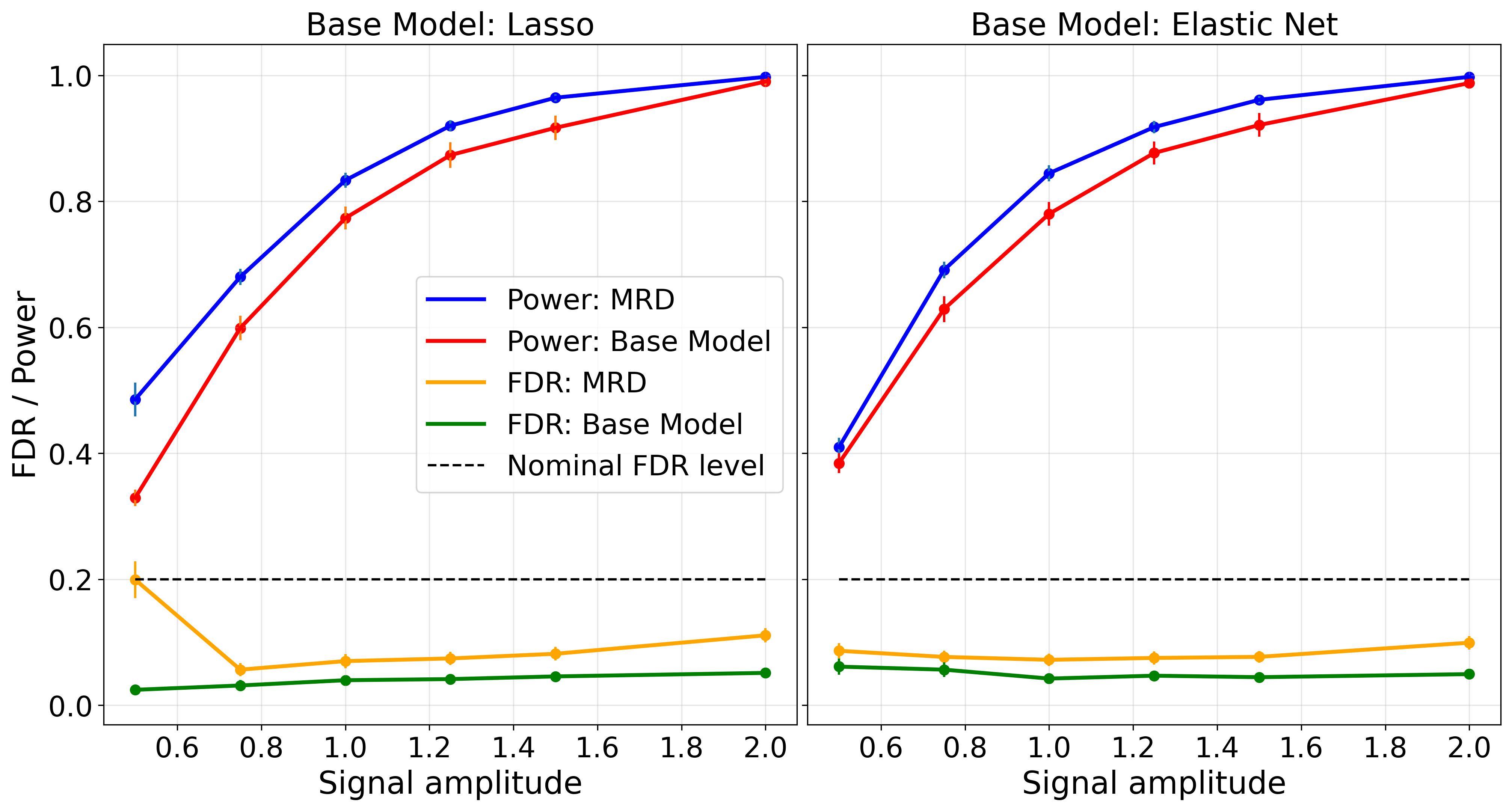}
    \caption{Semi-synthetic experiments with real human immunodeficiency virus mutation features and a simulated response that follows a linear model with varying signal strength $c$. Empirical power and FDR with $q= 0.2$ are evaluated over $50$ random train/test splits of the data. The other details are as in Figure~\ref{fig:non_lin_semi}.} 
    \label{fig:lin_semi}
\end{figure}

In addition to the non-linear semi-synthetic experiments described in Section~\ref{sec:real-data} of the main manuscript, in this section we perform variable selection with real human immunodeficiency virus mutation features and a simulated response that follows a \emph{linear} model with varying signal strength $c$. That is, we simulate $Y$ as described in \eqref{eq:y-mid-x} such that $y \mid X$ follows \textbf{M1}, while treating the real features $X$ as fixed. The results are presented in Figure~\ref{fig:lin_semi}. As demonstrated, the FDR obtained by all methods is below the target level $q=0.2$, indicating the robustness of the selection procedure to model misspecification. (Recall that we approximate the unknown $P_X$ as a multivariate Gaussian whose parameters are estimated from the whole data.) In terms of power, our MRD methods outperform the base predictive models. This is in line with the overall tendency of the experiments presented in this paper, and, in particular, with the ones the correspond to the non-linear semi-synthetic case.

\subsection{Semi-synthetic experiments with a varying number of samples}
\label{supp:semi_vary_n}

Herein, we extend the experiments from the previous sub-section and study the performance of our method as a function of the sample size. We follow the linear semi-synthetic setting described in Section~\ref{supp:semi_linear} with a fixed signal strength $c=3$, however sample data sets of different size by randomly choosing a subset of $n<1555$ observations from the whole data set. We apply the controlled variable selection procedure only on the selected subset, where the marginal distribution of $P_X$ is estimated using the entire data set; see section~\ref{sec:real-data}. The results obtained when choosing lasso to be the base predictive model are summarized in Table~\ref{tab:semi_vay_n}. As can be seen, the FDR is empirically controlled across all settings, and the power increases with the sample size. Also, consistent with previous experiments, the MRD approach outperforms the baseline method in terms of power.

\begin{table}[t]
    \caption{Semi-Synthetic experiments with real features and a simulated response that follows a linear model with varying number of samples $n$ and a fixed signal strength $c=3$. The empirical FDR (nominal level $q=0.2$) and power are evaluated by averaging over 50 independent experiments. All standard errors are below 0.02.}
     \label{tab:semi_vay_n}
\fontsize{8}{8}\selectfont
  \centering
  \ra{1}

  \begin{tabular}{cllllc}
    \toprule
    &\multicolumn{2}{c}{MRD Lasso}&
    \multicolumn{2}{c}{Lasso}&\multicolumn{1}{c}{\% imp. of}
    \\
    
    \cmidrule(r){2-3}
    \cmidrule(r){4-5}

    $n$ &Power     & FDR     & Power  & FDR & power  \\
    \midrule
    100 & 0.065 & 0.055 & 0.040 & 0.026&62.5 \\
    150 & 0.223 & 0.062 & 0.172 & 0.026&29.7 \\
    200 & 0.341 & 0.081 & 0.290 & 0.060&17.6 \\
    250 & 0.462 & 0.085 & 0.430 & 0.041& 7.4\\
    300 & 0.579 & 0.088 & 0.548 & 0.050& 5.7\\
    350 & 0.646 & 0.078 & 0.629 & 0.050& 2.7\\
    500 & 0.820 & 0.080 & 0.810 & 0.054& 1.2\\
    600 & 0.897 & 0.075 & 0.900 & 0.049& -0.3\\
    \bottomrule
  \end{tabular}
\end{table}

\end{document}